\documentclass{article}
   
\title{\huge {CaFA:} Global Weather Fore\textbf{Ca}sting \\ with \textbf{F}actorized \textbf{A}ttention on Sphere}

\usepackage{lipsum}
\usepackage[utf8]{inputenc}
\usepackage[T1]{fontenc}    
\usepackage[bookmarks=true, colorlinks=true,linkcolor=blue!70,citecolor=green!80!black]{hyperref}
\usepackage[round]{natbib}
\usepackage{wrapfig}
\usepackage{booktabs}  
\usepackage{longtable}
\usepackage{ducksay}
\usepackage{multicol}
\usepackage{etaremune}
\usepackage{afterpage}
\usepackage{capt-of}
\usepackage[table, x11names]{xcolor}
\usepackage{decorule}
\usepackage{scalerel,xparse}
\usepackage{enumitem} 
\usepackage[top=25mm, bottom=25mm, left=25mm, right=25mm,
            foot=5mm, marginparsep=0mm]{geometry}
\usepackage{tabularray}
\usepackage{multirow}
\usepackage{anyfontsize}
\usepackage{subcaption}
\usepackage[symbol*]{footmisc}

\usepackage{chngcntr}
\counterwithin{figure}{section} 
\counterwithin{table}{section}
\usepackage{graphicx}
\usepackage{tikz}
\usepackage{tikz-cd}
\usepackage{hf-tikz}
\usepackage{pgfplots} 
\pgfplotsset{compat=newest} 
\pgfplotsset{
        table/search path={figures/drawings},
    }
\usetikzlibrary{fadings}
\usetikzlibrary{shapes, arrows, fit, backgrounds, arrows.meta}

\usetikzlibrary{matrix}
\usetikzlibrary{shadows.blur}
\usetikzlibrary{patterns, tikzmark}
\usetikzlibrary{decorations.pathreplacing, calc, decorations.markings,}
\usetikzlibrary{positioning}
\usetikzlibrary{shapes.geometric}
\pgfdeclareplotmark{mystar}{
    \node[star,star point ratio=2.25,minimum size=5pt,
          inner sep=0pt,draw=black,solid,fill=green] {};
}

\usepgfplotslibrary{groupplots}
\usepgfplotslibrary{patchplots}

\definecolor{bg}{gray}{0.97}
\definecolor{olive}{rgb}{0.6, 0.6, 0.2}
\definecolor{sand}{rgb}{0.8666666666666667, 0.8, 0.4666666666666667}
\definecolor{wine}{rgb}{0.5333333333333333, 0.13333333333333333, 0.3333333333333333}
\definecolor{deblue}{RGB}{11,132,147}
\definecolor{ocra}{RGB}{204, 119, 34}

\usepackage{lettrine}
\usepackage{Typocaps}

\LettrineTextFont{\itshape}
\usepackage{amsthm}

\usepackage{amssymb}
\usepackage{pifont}
\usepackage{minitoc}

\newcommand{\chapref}[1]{\hyperref[#1]{Chapter \ref{#1}}}
\newcommand{\secref}[1]{\hyperref[#1]{Section \ref{#1}}}

\usepackage{marginnote}

\usepackage[many]{tcolorbox}
\tcbuselibrary{breakable,xparse,skins}

\usepackage{newunicodechar}
\newunicodechar{λ}{{$\mathtt\lambda$}}
\newunicodechar{μ}{{$\mathtt\mu$}}

\usepackage{xparse}
\DeclareTColorBox{emphbox}{O{black}O{0cm}}{
    enhanced jigsaw,
    breakable,
    outer arc=0pt,
    arc=0pt,
    colback=white,
    rightrule=0pt,
    toprule=0pt,
    top=0pt,
    right=0pt,
    bottom=0pt,
    bottomrule=0pt,
    colframe=#1,
    enlarge left by=#2,
    width=\linewidth-#2,
}

\DeclareTColorBox{emphbox}{O{black}O{0cm}}{
    empty,
    breakable=true,
    outer arc=0pt,
    arc=0pt,
    rightrule=0pt,
    leftrule=2pt,
    borderline west={2pt}{0pt}{#1},
    toprule=0pt,
    top=0pt,
    right=-3pt,
    bottom=0pt,
    bottomrule=0pt,
    colframe=#1,
    enlarge left by=#2,
    width=\linewidth-#2,
}

\RequirePackage{algorithm}
\RequirePackage{algorithmic}

\usepackage{tikz}
\usepackage{collcell}
\usepackage{nicematrix}

\definecolor{high}{HTML}{1034A6} 
\definecolor{low}{HTML}{F0F8FF}

\newcommand{\gradientcell}[6]{
    \ifdimcomp{#1pt}{>}{#3 pt}{#1}{
    \ifdimcomp{#1pt}{<}{#2 pt}{#1}{
         \pgfmathparse{int(round(100*((#1-#2)/(#3-#2))))))}%
        \xdef\tempa{\pgfmathresult}%
        \cellcolor{#5!\tempa!#4!#6}{#1}%
    }}%
 }
 \newcommand{\rmsepressureone}[1]{
 \gradientcell{#1}{39}{44}{high}{low}{50}%
 }

  \newcommand{\rmsepressurethree}[1]{
 \gradientcell{#1}{123}{135}{high}{low}{50}%
 }

  \newcommand{\rmsepressurefive}[1]{
 \gradientcell{#1}{268}{304}{high}{low}{50}%
 }

  \newcommand{\rmsepressureseven}[1]{
 \gradientcell{#1}{433}{521}{high}{low}{50}%
 }

\newcommand{\rmsetemperatureone}[1]{
 \gradientcell{#1}{0.51}{0.63}{high}{low}{50}%
 }

 \newcommand{\rmsetemperaturethree}[1]{
 \gradientcell{#1}{0.93}{1.15}{high}{low}{50}%
 }

 \newcommand{\rmsetemperaturefive}[1]{
 \gradientcell{#1}{1.53}{1.79}{high}{low}{50}%
 }

 \newcommand{\rmsetemperatureseven}[1]{
 \gradientcell{#1}{2.11}{2.59}{high}{low}{50}%
 }

 \newcommand{\rmsetsurfaceone}[1]{
 \gradientcell{#1}{0.51}{0.56}{high}{low}{50}%
 }
  \newcommand{\rmsetsurfacethree}[1]{
 \gradientcell{#1}{0.82}{0.91}{high}{low}{50}%
 }
  \newcommand{\rmsetsurfacefive}[1]{
 \gradientcell{#1}{1.20}{1.39}{high}{low}{50}%
 }
 \newcommand{\rmsetsurfaceseven}[1]{
 \gradientcell{#1}{1.58}{1.97}{high}{low}{50}%
 }
 
\newcommand{\rmsepressuresurfaceone}[1]{
 \gradientcell{#1}{48}{63}{high}{low}{50}%
 }
 \newcommand{\rmsepressuresurfacethree}[1]{
 \gradientcell{#1}{130}{150}{high}{low}{50}%
 }
 \newcommand{\rmsepressuresurfacefive}[1]{
 \gradientcell{#1}{270}{310}{high}{low}{50}%
 }
 \newcommand{\rmsepressuresurfaceseven}[1]{
 \gradientcell{#1}{415}{508}{high}{low}{50}%
 }

 \newcommand{\rmsehumidityone}[1]{
 \gradientcell{#1}{0.47}{0.56}{high}{low}{50}%
 }
 \newcommand{\rmsehumiditythree}[1]{
 \gradientcell{#1}{0.80}{0.97}{high}{low}{50}%
 }
 \newcommand{\rmsehumidityfive}[1]{
 \gradientcell{#1}{1.06}{1.28}{high}{low}{50}%
 }
 \newcommand{\rmsehumidityseven}[1]{
 \gradientcell{#1}{1.22}{1.54}{high}{low}{50}%
 }%
  \newcommand{\rmseuwindone}[1]{
 \gradientcell{#1}{1.0}{1.18}{high}{low}{50}%
 }
 \newcommand{\rmseuwindthree}[1]{
 \gradientcell{#1}{1.93}{2.32}{high}{low}{50}%
 }
 \newcommand{\rmseuwindfive}[1]{
 \gradientcell{#1}{3.09}{3.65}{high}{low}{50}%
 }
 \newcommand{\rmseuwindseven}[1]{
 \gradientcell{#1}{3.96}{4.99}{high}{low}{50}%
 }
 
\newcommand{\rmsevwindone}[1]{
 \gradientcell{#1}{1.02}{1.20}{high}{low}{50}%
 }
 \newcommand{\rmsevwindthree}[1]{
 \gradientcell{#1}{1.93}{2.34}{high}{low}{50}%
 }
 \newcommand{\rmsevwindfive}[1]{
 \gradientcell{#1}{3.13}{3.69}{high}{low}{50}%
 }
 \newcommand{\rmsevwindseven}[1]{
 \gradientcell{#1}{4.03}{5.07}{high}{low}{50}%
 }

  \newcommand{\rmseusurfaceone}[1]{
 \gradientcell{#1}{0.65}{0.83}{high}{low}{50}%
 }
 \newcommand{\rmseusurfacethree}[1]{
 \gradientcell{#1}{1.32}{1.61}{high}{low}{50}%
 }
 \newcommand{\rmseusurfacefive}[1]{
 \gradientcell{#1}{2.16}{2.56}{high}{low}{50}%
 }
 \newcommand{\rmseusurfaceseven}[1]{
 \gradientcell{#1}{2.74}{3.50}{high}{low}{50}%
 }

  \newcommand{\rmsevsurfaceone}[1]{
 \gradientcell{#1}{0.68}{0.87}{high}{low}{50}%
 }
 \newcommand{\rmsevsurfacethree}[1]{
 \gradientcell{#1}{1.37}{1.67}{high}{low}{50}%
 }
 \newcommand{\rmsevsurfacefive}[1]{
 \gradientcell{#1}{2.24}{2.66}{high}{low}{50}%
 }
 \newcommand{\rmsevsurfaceseven}[1]{
 \gradientcell{#1}{2.87}{3.66}{high}{low}{50}%
 }

 \newcommand{\bcpressureone}[1]{
 \gradientcell{#1}{51}{96}{high}{low}{50}%
 }

  \newcommand{\bcpressurethree}[1]{
 \gradientcell{#1}{170}{244}{high}{low}{50}%
 }

  \newcommand{\bcpressurefive}[1]{
 \gradientcell{#1}{348}{440}{high}{low}{50}%
 }

  \newcommand{\bcpressureseven}[1]{
 \gradientcell{#1}{544}{599}{high}{low}{50}%
 }

  \newcommand{\bctempone}[1]{
 \gradientcell{#1}{0.59}{1.11}{high}{low}{50}%
 }

  \newcommand{\bctempthree}[1]{
 \gradientcell{#1}{1.04}{1.59}{high}{low}{50}%
 }

  \newcommand{\bctempfive}[1]{
 \gradientcell{#1}{1.71}{2.23}{high}{low}{50}%
 }

  \newcommand{\bctempseven}[1]{
 \gradientcell{#1}{2.47}{2.77}{high}{low}{50}%
 }

  \newcommand{\bctempsurfone}[1]{
 \gradientcell{#1}{0.49}{1.10}{high}{low}{50}%
 }

  \newcommand{\bctempsurfthree}[1]{
 \gradientcell{#1}{0.82}{1.43}{high}{low}{50}%
 }

  \newcommand{\bctempsurffive}[1]{
 \gradientcell{#1}{1.28}{1.83}{high}{low}{50}%
 }

  \newcommand{\bctempsurfseven}[1]{
 \gradientcell{#1}{1.79}{2.18}{high}{low}{50}%
 }

   \newcommand{\bcusurfone}[1]{
 \gradientcell{#1}{0.58}{1.41}{high}{low}{50}%
 }

  \newcommand{\bcusurfthree}[1]{
 \gradientcell{#1}{1.28}{2.18}{high}{low}{50}%
 }

  \newcommand{\bcusurffive}[1]{
 \gradientcell{#1}{2.15}{2.94}{high}{low}{50}%
 }

  \newcommand{\bcusurfseven}[1]{
 \gradientcell{#1}{2.97}{3.43}{high}{low}{50}%
 }

\newcommand{\bcflops}[1]{
 \gradientcell{#1}{0.61}{2.22}{high}{low}{50}%
 }

\usepackage{amsmath, amssymb, amsfonts, mathtools}
\usepackage{physics}
\usepackage{cancel}
\usepackage{thmtools, thm-restate}
\usepackage{accents}
\usepackage{bm}

\makeatletter
\newcommand{\ostar}{\mathbin{\mathpalette\make@circled *}}
\newcommand{\make@circled}[2]{%
  \ooalign{$\m@th#1\smallbigcirc{#1}$\cr\hidewidth$\m@th#1#2$\hidewidth\cr}%
}
\newcommand{\smallbigcirc}[1]{%
  \vcenter{\hbox{\scalebox{0.77778}{$\m@th#1\bigcirc$}}}%
}
\makeatother

\usepackage{pict2e, picture}
\makeatletter
\DeclareRobustCommand{\Arrow}[1][]{%
\check@mathfonts
\if\relax\detokenize{#1}\relax
\settowidth{\dimen@}{$\m@th\rightarrow$}%
\else
\setlength{\dimen@}{#1}%
\fi
\sbox\z@{\usefont{U}{lasy}{m}{n}\symbol{41}}%
\begin{picture}(\dimen@,\ht\z@)
\roundcap
\put(\dimexpr\dimen@-.7\wd\z@,0){\usebox\z@}
\put(0,\fontdimen22\textfont2){\line(1,0){\dimen@}}
\end{picture}%
}
\makeatother

\DeclareMathAlphabet{\nummathbb}{U}{BOONDOX-ds}{m}{n}

\makeatletter
\DeclareRobustCommand\widecheck[1]{{\mathpalette\@widecheck{#1}}}
\def\@widecheck#1#2{%
    \setbox\z@\hbox{\m@th$#1#2$}%
    \setbox\tw@\hbox{\m@th$#1%
       \widehat{%
          \vrule\@width\z@\@height\ht\z@
          \vrule\@height\z@\@width\wd\z@}$}%
    \dp\tw@-\ht\z@
    \@tempdima\ht\z@ \advance\@tempdima2\ht\tw@ \divide\@tempdima\thr@@
    \setbox\tw@\hbox{%
       \raise\@tempdima\hbox{\scalebox{1}[-1]{\lower\@tempdima\box
\tw@}}}%
    {\ooalign{\box\tw@ \cr \box\z@}}}
\makeatother

\begin{document}
 \author{%
  Zijie Li$^{\dagger}$, Anthony Zhou$^\dagger$, Saurabh Patil$^\dagger$, Amir Barati Farimani$^{\dagger}$\footnote{Correspondence: \texttt{barati@cmu.edu}}\\ 
  \textit{$^\dagger$ Mechanical Engineering Department, Carnegie Mellon University}}
\maketitle
 \begin{abstract}
Accurate weather forecasting is crucial in various sectors, impacting decision-making processes and societal events. Data-driven approaches based on machine learning models have recently emerged as a promising alternative to numerical weather prediction models given their potential to capture physics of different scales from historical data and the significantly lower computational cost during the prediction stage. Renowned for its state-of-the-art performance across diverse domains, the Transformer model has also gained popularity in machine learning weather prediction. Yet applying Transformer architectures to weather forecasting, particularly on a global scale is computationally challenging due to the quadratic complexity of attention and the quadratic increase in spatial points as resolution increases. In this work, we propose a factorized-attention-based model tailored for spherical geometries to mitigate this issue.  More specifically, it utilizes multi-dimensional factorized kernels that convolve over different axes where the computational complexity of the kernel is only quadratic to the axial resolution instead of overall resolution. The deterministic forecasting accuracy of the proposed model on $1.5^\circ$ and 0-7 days' lead time is on par with state-of-the-art purely data-driven machine learning weather prediction models.  We also showcase the proposed model holds great potential to push forward the Pareto front of accuracy-efficiency for Transformer weather models, where it can achieve better accuracy with less computational cost compared to Transformer based models with standard attention.
\end{abstract}

\setlength\abovedisplayshortskip{2pt}
\setlength\belowdisplayshortskip{2pt}
\setlength\abovedisplayskip{2pt}
\setlength\belowdisplayskip{2pt}

\section{Introduction}

Weather forecasting plays a pivotal role in numerous aspects of modern society, spanning from agriculture and transportation to disaster preparedness and public safety. The ability to predict weather conditions accurately allows informed decision making in a wide array of engineering and social events. Numerical weather prediction (NWP) \citep{NWPreview2015Nature}, a cornerstone of modern meteorology, utilizes numerical models to simulate the Earth's atmosphere and predict the future weather. The simulation process consists of constructing partial differential equations to describe the global weather dynamics and then solving the equation via numerical algorithms. Resolving finer physical scales, such as simulating microphysics phenomena or using a finer spatio-temporal discretization, allows greater NWP accuracy; however, this greatly increases the computational cost. Performing a 10-day forecast can take an hour on a supercomputer \citep{ecmwf-progress, nwp-progress}. Additionally, accurately resolving small scale physics requires a well-tailored parametrization of the numerical model that approximates the underlying physical process. 

The recent progress in machine learning and increasing amount of available data have facilitated the emergence of machine-learning-based weather prediction (MLWP) models. Compared to NWP, MLWP offers the potential to capture and learn finer-scale physics from data without describing the physics in an analytical mathematical model or searching for precise parameterization fits. By ingesting datasets encompassing large amounts of historical weather records, these models can learn and fit intricate patterns and correlations in the global weather dynamics. 
Recently, MLWP models, particularly those based on deep neural networks, have shown promising potential in global weather prediction \citep{pangu2023nature, graphcast2023science, kurth2023fourcastnet, chen2023fuxi, chen2023fengwu, climax2023icml, nguyen2023stormer, verma2024climode, sfno2023icml, keisler2022forecasting}. They have demonstrated competitive accuracy compared to physics-based numerical models in medium-range weather forecasting. Once trained, they can conduct prediction tasks very efficiently on computing devices (e.g. graphic processing units) specialized for neural network computation. For instance, a 10-day forecast at a scale of around 30 km can be completed within a minute \citep{pangu2023nature, graphcast2023science, kurth2023fourcastnet}.

Among different types of neural network architectures, Transformer models based on the attention mechanism \citep{attention2017nips} stands out as a notable class. Initially designed for processing sequences in natural language, Transformer models have found successful applications in a wide array of engineering and scientific domains, such as image classification \citep{vit2021iclr} and protein folding \citep{alphafold2021nature}.  
At its essence, the Transformer's attention mechanism provides the ability to capture non-local patterns at all length scales within the context and employs data-controlled weights to aggregate different features dynamically, with its computational cost scaling quadratically with respect to the context size; this can range from one to the full length of the input data. Prior efforts to reduce the computational cost include linearizing the computational complexity via kernel approximation \citep{performer2020iclr}, hardware-aware optimization of the algorithm implementation \citep{flashattention2022nips}, or sparsifying the attention computation by restricting the context size through partitioning the input into different windows \citep{beltagy2020longformer, swin2021cvpr}. For global weather forecasting tasks, previous Transformer-based models apply attention to the input mostly with the spatial structure flattened \citep{climax2023icml, nguyen2023stormer}. Sliding window attention can be used afterwards to restrict the attention context \citep{chen2023fuxi, pangu2023nature, chen2023fengwu} and therefore reduce the computational cost. In these cases, the multi-dimensional (longitude-latitude) spatial structure of the input is not taken into account explicitly. While attending between every pair of tokens that represent embedded features at different locations can exploit all possible interactions in the data, its computation is prohibitively expensive. Another drawback of discarding spatial structure is that useful inductive biases about the underlying physics such as boundary condition and geometry information (the target domain is on a sphere) can not be easily applied, although some of this information can be implicitly encoded to the model, e.g. via positional encoding.

In this work, we explore a different perspective of designing the attention layer, where the multi-dimensional structure is preserved during the attention calculation. We build our model with the low-rank attention kernel operator \citep{factformer2023nips} that decomposes the multi-dimensional attention kernel into a set of single-dimensional sub-kernels. The sub-kernels are parameterized using dot-product attention with angular distance modulation which facilitates a smooth response of the attention score under varying distances. We numerically validate the deterministic forecasting capability of the proposed model on WeatherBench2 \citep{rasp2023weatherbench2}'s data which has a resolution of $1.5^\circ$. The proposed model - CaFA (Fore\textbf{Ca}st with \textbf{F}actorized \textbf{A}ttention) outperforms the deterministic operational Integrated Forecast System (IFS) on most investigated variables and shows competitive performance over other state-of-the-art end-to-end MLWPs in 0-7 days' global weather forecasting. Our study represents a stride towards reducing the computational cost of Transformer-based weather models without sacrificing accuracy, thereby enhancing their accessibility and efficiency.

\section{Related Works}

The advances in computing and general numerical methods for partial differential equations have revolutionized numerical weather prediction over the last century. The core of numerical modeling of weather lies in describing the complex Earth dynamics including land, sea and atmosphere, as  coupled equation systems, or earth system models (ESM) \citep{EarthSystemModel}. These equation systems are often simulated via general circulation models (GCM) \citep{philips-gcm, lynch-weather-prediction, satoh2013atmospheric} numerically. Under-resolved physics in GCMs (sub-grid physics that are beyond the scale of finest discretization) are usually approximated with empirical parameterization schemes \citep{Stensrud2007parametrization} and adjusted with domain expertise \citep{sherwood2014spread, webb2013origins}. Not only have the numerical methods of climate/weather prediction been greatly advanced, the amount of high-quality observed and analyzed data has also increased vastly over the last century. One of the most notable instances is historical data from the European Center for Medium-Range Weather Forecasts (ECMWF) reanalysis v5 (ERA5) \citep{ERA52020}, which combines historical observations with results from high-fidelity integrated Forecasting System (IFS) \citep{wedi2015ifs}. These datasets offer an alternative for advancements in climate and weather modeling, that is using data-driven models instead of manually adjusting parameterizations to better capture underlying physics. Various machine learning models have been employed to enhance the predictive capabilities of complex atmospheric processes, including earlier models that target specific variables/regions \citep{david2014mlstorm, Lagerquist2017} or global weather variables at relatively coarser resolutions \citep{Dueben2018, scher2018, scher2019, weyn2019, rasp2021resnet, weyn2020improving}.  More recently, \citet{keisler2022forecasting} proposes a graph-neural-network-based model for forecasting multiple surface and upper-air variables on $1^{\circ}$ resolution with accuracy surpassing many prior MLWPs, which underscores the vast potential of MLWP in enhancing weather prediction accuracy and efficiency. FourCastNet \citep{kurth2023fourcastnet} demonstrates the promising forecasting capability and superior inference efficiency of MLWP on $0.25^{\circ}$ resolution. Pangu-Weather \citep{pangu2023nature} and then GraphCast \citep{graphcast2023science} further elevate the accuracy of MLWP to the level such that their medium-range forecast (within 10 days) surpasses the state-of-the-art deterministic NWP on many weather variables. They marked the first milestone where MLWP can outperform NWP in terms of both deterministic forecasting accuracy and inference efficiency at such a high resolution.

Each of these advancements are built upon a specific neural network architecture. FourCastNet, which is built upon Adaptive Fourier Neural Operator \citep{guibas2022afno, li2021fno}, employs spectral convolution in the Fourier space to capture interactions in the spatial domain. Followup work with the Spherical Fourier Neural Operator \citep{sfno2023icml} improves and extends this framework by doing the convolution in spherical harmonic space to better account for the underlying spherical geometry. GraphCast scales up the graph neural network in \citet{keisler2022forecasting} and leverages multi-scale message passing to better capture both local and non-local interaction between different spatial points. Pangu-Weather is based on the Swin-Transformer \citep{swin2021cvpr} that leverages sliding window attention to capture spatial interactions. FuXi \citep{chen2023fuxi} and Fengwu \citep{chen2023fengwu}, which are also based on the Swin Transformer, improves the model's long-term forecasting accuracy with an improved training strategy and model architectures. In parallel, ClimaX \citep{climax2023icml} proposes a versatile climate modelling framework with a modified Vision Transformer (ViT) \citep{vit2021iclr} backbone and showcases it's ability to address various climate and weather tasks including global weather prediction. Stormer \citep{nguyen2023stormer} scales up ClimaX's Transformer architecture and proposes to pretrain the model with dynamic lead times to further enhance the long-term forecasting accuracy.

Beyond deterministic regression, the integration of deep-learning-based generative models into weather modeling has also gained traction, including ensemble prediction \citep{price2023gencast, li2023seeds}, probabilistic forecasting and nowcast \citep{leinonen2023latent, sonderby2020metnet, nath2023forecasting} and downscaling \citep{leinonen2020sr, mardani2023cordiff, chen2023swinrdm}.  Moreover, machine learning models can also be combined with physics priors to create a hybrid model, which can potentially further improve the robustness and accuracy. A canonical paradigm is using machine learning model to predict the under-resolved processes \citep{rasp2018subgrid} such as various subgrid physics while using a numerical solver built upon governing equation to solve the overall dynamics propagation. NeuralGCM \citep{kochkov2023neural} presents a pioneering realization of this paradigm with performance on par with state-of-the-art MLWP and NWP in terms of both deterministic and ensemble forecast. Additionally, physics can be incorporated by formulating the system propagation as a specific ordinary differential equation \citep{verma2024climode} to guarantee certain physics constraints like momentum conservation.

\newpage
\begin{figure}[H]
    \centering
    \includegraphics[width=\linewidth]{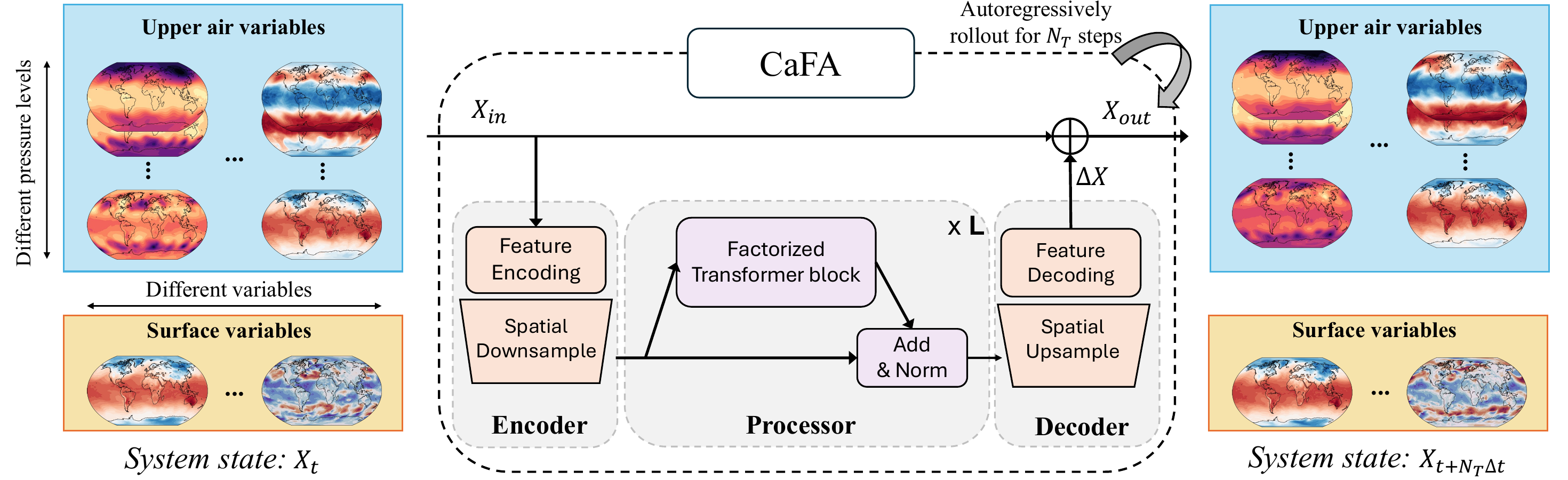}

    \caption{Main schematic of the proposed Transformer-based weather forecast model - CaFA. The model approximates a Markovian mapping that forwards the last system state to the next system state with a fixed time interval $\Delta t$.}
    \label{fig:main-scheme}
    \vspace{+4mm}
    \includegraphics[width=\linewidth]{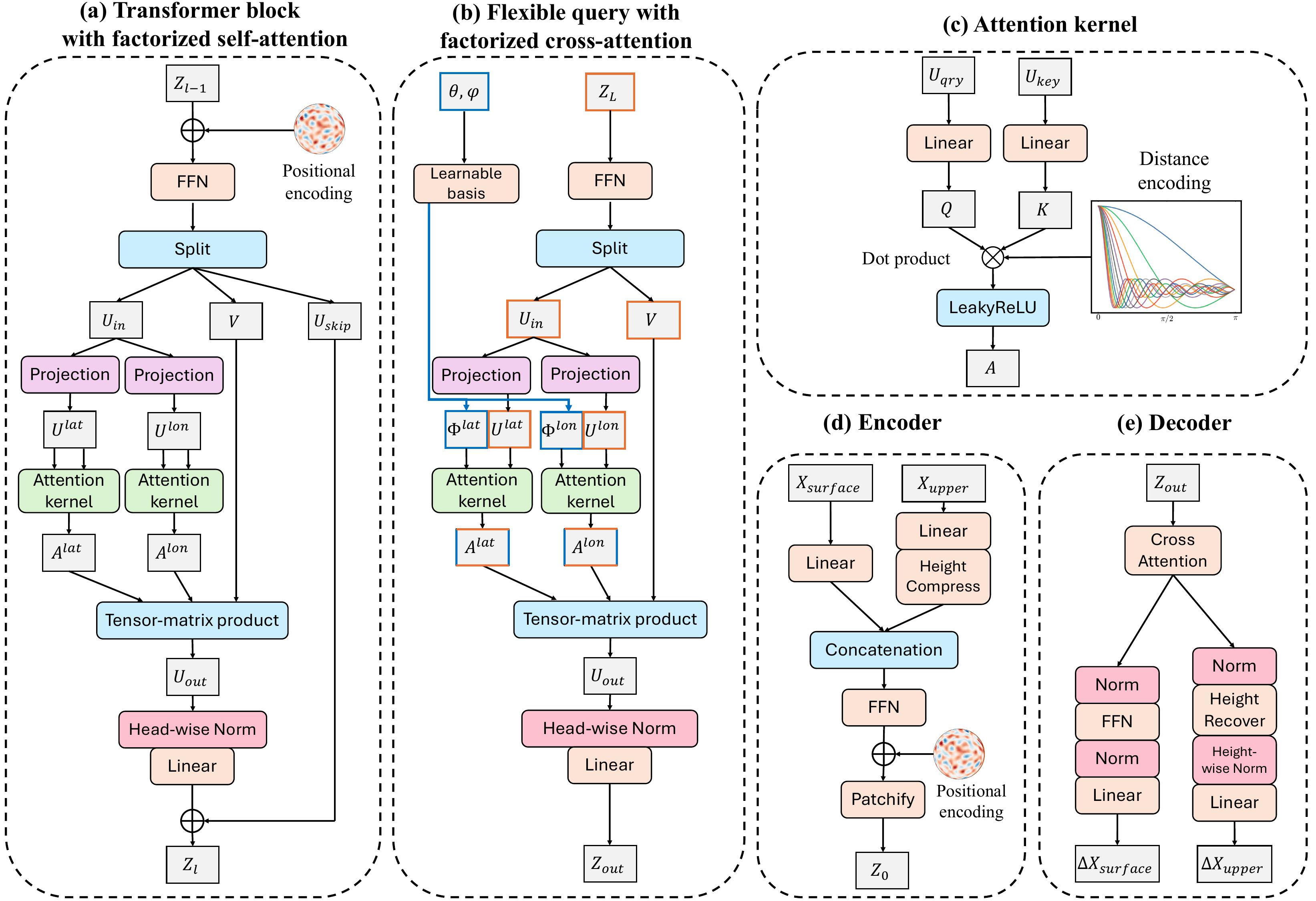}
    \caption{Main architecture of the model. \textbf{(a)} Each Transformer block in the Processor comprises of a feed-forward network (abbreviated as "FFN", which is a two-layer MLP) and a factorized attention layer. The factorized attention layer contains three major components: projection, attention and tensor-matrix product. \textbf{(b)} Cross-attention is used to project the latent embedding on a coarse grid to the high-resolution grid (i.e. upsample). The query location (at longitude $\theta$ and latitude $\varphi$) can be arbitrary points in the domain. \textbf{(c)} Dot-product attention with relative distance encoding. \textbf{(d)} The Encoder encodes the input surface variables and upper-air variables into latent embeddings and projects them to a coarser latent grid via patchifying. \textbf{(e)} The Decoder projects the latent grid back to the original grid and decodes the latent embedding into surface and upper-air variables' residual.}
    \label{fig:main-arch}
\end{figure}

\newpage

\section{Methodology}

\subsection{Attention mechanism}
The attention mechanism \citep{attention2014iclr, attention2017nips} has become the core component of modern neural network architectures, revolutionizing various fields including natural language processing \citep{bert2019acl, gpt2020nips} and computer vision \citep{vit2021iclr}. Attention operates on three sets of vectors: queries $\{\mathbf{q}_i\}_{i=1}^{N_q}$, keys $\{\mathbf{k}_i\}_{i=1}^{N_k}$, and values $\{\mathbf{v}_i\}_{i=1}^{N_v}$ ($N_k=N_v$) and computes the weighted average of values:
\begin{equation}
\label{eq:attn computation}
    \mathbf{z}_i
    = \sum_{j=1}^{N_v}h(\mathbf{q}_i, \mathbf{k}_j)\mathbf{v}_j,
\end{equation}
where $\mathbf{q}_i, \mathbf{k}_i, \mathbf{v}_i \in \mathbb{R}^{d}$,  $h: \mathbb{R}^{d} \times \mathbb{R}^{d} \mapsto \mathbb{R}$ is a suitable weight function such as softmax with a scaling factor \citep{attention2017nips}: $h(\mathbf{q}_i,  \mathbf{k}_j) = \exp(\mathbf{q}_i \cdot\mathbf{k}_j/\tau) / \sum_{s}\exp(\mathbf{q}_i \cdot\mathbf{k}_s/\tau)$ with scaling $\tau=\sqrt{d}$. For self-attention, the query/key/value is transformed from the same input $\mathbf{u}_i$ through a learnable linear transform: 
\begin{equation}
\label{eq:attn projection}
    \mathbf{q}_i=\mathbf{u}_i W_q, \mathbf{k}_i=\mathbf{u}_i W_k, \mathbf{v}_i=\mathbf{u}_i W_v, 
\end{equation}
where $\mathbf{u}_i \in \mathbb{R}^{ d_{\text{in}}}$ is the input vector and $W_q, W_k, W_v \in \mathbb{R}^{d_{\text{in}}\times d}$ are learnable weight matrices. In cross-attention, queries are usually obtained from a different input source from the keys and values. In its continuous limit, attention in Equation \eqref{eq:attn computation} parametrizes a kernel integral transform operator \citep{kovachki2023neural, galerkin2021nips}. With suitable mesh weights it becomes a Riemann sum approximation of the kernel integral transform:
\begin{equation}
  \label{eq:attn kernel}
  \mathbf{z}_i=\sum_{j=1}^{N_v}h(\mathbf{q}_i \cdot \mathbf{k}_j) \mathbf{v}_j \mu_j \approx \int_{\Omega} \kappa\left(x_i, \xi\right)v(\xi) d\xi,
\end{equation}
where $\mu_j$ is the mesh-based weight for grid point $x_j$, and the weight function $h$ together with dot product of $\mathbf{q}_i, \mathbf{k}_j$ paramterizes the kernel function $\kappa$. The application of the attention in modeling a diverse range of PDE-related problems \citep{galerkin2021nips, bvptransformer2023iclr, oformer2023tmlr, factformer2023nips, gnot2023icml, liu2024hano, romer2023, loca2022jmlr} suggests its potential for learning the complex global weather dynamics. It is also observed that in \citet{factformer2023nips} the attention matrices of a Transformer trained on PDE problems often have a low-rank structure which motivates us to explore a low-rank parametrization. 

\subsection{Axial factorized attention on sphere}
\paragraph{Factorized attention} The original formulation of attention has a complexity $O(N^2)$ (with respect to the total number of grid points $N$) in terms of its floating point operation. As the number of grid points grows exponentially with respect to the dimension of the problem, the computational cost of applying attention to higher dimensional problems is relatively expensive. To mitigate this issue, \citet{factformer2023nips} proposes an axial-factorized attention leveraging the tensorized structure of data, which comprises two learnable integral operators: axial projection and axial factorized attention. The projection operator projects input functions with a high-dimensional domain $\Omega$ into $n$ sub-functions with a single-dimensional domain ($\Omega = \Omega_1 \times \Omega_2 \hdots \times \Omega_n$), and then the attention kernels are computed based on projected single-dimensional functions. Below we will introduce their discretized forms.

The projection operator is defined as: 
\begin{equation}
     \text{\textit{Projection}:} \quad U^{(m)}_{i^{(m)}} = \gamma \left( \sum_{j^{(1)}=1}^{S_1} \mu^{(1)}_{j^{(1)}}  \hdots 
     \sum_{j^{(m-1)}}^{S_{m-1}} \mu^{(m-1)}_{j^{(m-1)}}
     \sum_{j^{(m+1)}}^{S_{m+1}} \mu^{(m+1)}_{j^{(m+1)}} \hdots
     \sum_{j^{(n)}}^{S_n} \mu^{(n)}_{j^{(n)}} U_{j^{(1)} j^{(2)}\hdots i^{(m)} \hdots j^{(n)}}
     \right), \label{general projection} 
\end{equation}
where $U^{(m)} \in \mathbb{R}^{S_m \times d}$ is the projected feature along the $m$-th axis, $d$ is the number of feature channels, 
$U$ is obtained from input tensor $Z \in \mathbb{R}^{S_1 \times \hdots \times S_n\times d}$ via a pointwise linear transformation, 
$S_m$ denotes the size of the grid that discretizes sub-domain $\Omega_m$ with $\mu^{(m)}_{j^{(m)}}$ being the corresponding mesh weights for grid point $j^{(m)}$, and $\gamma: \mathbb{R}^d \mapsto \mathbb{R}^d$ is a pointwise two-layer multi-layer perceptron. 

Based on the projected functions with a single-dimensional domain, the (self) axial factorized attention is computed as follows: 
\begin{align}
    \text{\textit{Axial attention kernel}:} &\quad A^{(m)} = Q^{(m)}(K^{(m)})^T, \text{where: } Q^{(m)} = U^{(m)}W_q^{(m)}, K^{(m)}=U^{(m)} W_k^{(m)}, \label{general axial attention} \\
     \text{\textit{Tensor-matrix product}:} &\quad \left( Z_{\text{out}} \right)_{i^{(1)} i^{(2)}\hdots i^{(n)}} \notag\\
     = &\sum_{i^{(1)}=1}^{S_1} \mu^{(1)}_{j^{(1)}} A^{(1)}_{i^{(1)}j^{(1)}}\sum_{i^{(2)}=1}^{S_2} \mu^{(2)}_{j^{(2)}}A^{(2)}_{i^{(2)}j^{(2)}} \hdots \sum_{i^{(n)}=1}^{S_n} \mu^{(n)}_{j^{(n)}} A^{(n)}_{i^{(n)}j^{(n)}} ~ V_{j^{(1)} j^{(2)}\hdots j^{(n)}}, \label{general tensor-matrix product}
\end{align} 
where $A^{(m)} \in \mathbb{R}^{S_m \times S_m}$ is the $m$-th axial kernel computed with dot product attention,  $V$ is the value tensor which is derived from the input via a learnable linear transformation following standard attention (Equation \eqref{eq:attn projection}),  and $W_q^{(m)}, W_k^{(m)}$ are learnable linear transformation matrices for computing queries and keys. In \citet{factformer2023nips}, all the problems considered have a uniform equi-spaced grid, thus the mesh weight on each grid point along $m$-th axis is $\mu^{(m)}_j=1/S_m,\forall j \in \{1,2,\hdots,S_m\}$ and the weighted sum in Equation \eqref{general projection} simply amounts to average pooling over all but the $m$-th axis. With an equi-angular grid on the spherical domain $\mathbb{S}^2$ and assuming uniform quadrature rule, the mesh weight on each latitude grid point is non-uniform. Given a latitude coordinate $\varphi \in [-\pi/2, \pi/2]$ and longitude coordinate $\theta \in [0, 2\pi)$ (and assuming unit radius), the integration of function $f: \mathbb{S}^2 \mapsto \mathbb{R}$ on the sphere takes the form:
\begin{equation}
   \int_{\mathbb{S}^2} f dS = 
   \int_{0}^{2\pi}  \int_{-\pi/2}^{\pi/2} f \cos \varphi  ~d\varphi d\theta,  
\end{equation}
where a suitable mesh weight for each latitude grid point (cell) $j$ is $\mu^{(\text{lat})}_j =  \frac{\pi}{N_{\text{lat}}}\cos \varphi_j$ and $\mu^{(\text{lon})}_j =  \frac{2\pi}{N_{\text{lon}}}$ for a longitude grid point on an equi-angular grid. In summary, the vanilla attention-based kernel integral in Equation \eqref{eq:attn kernel} has now been modified to:
\begin{equation}
    \label{eq:spherical attn kernel}
   z(\theta_i, \varphi_i)=\int_{0}^{2\pi} \kappa_{\text{lon}}(\theta_i, \theta) \int_{-\pi/2}^{\pi/2} \kappa_{\text{lat}}(\varphi_i, \varphi) v(\varphi) \cos \varphi  ~d\varphi d\theta,
\end{equation}
with latitudinal kernel and longitudinal kernel $\kappa_{\text{lon}}, \kappa_{\text{lat}}$ parameterized using dot product attention with angular distance modulation which will be described in the next paragraph. Illustrative figures comparing factorized attention and standard attention are provided in Appendix Figure \ref{fig:spherical pe}, \ref{fig:attention kernel visualization}, \ref{fig:distance encoding visualization}.

\paragraph{Distance encoding and positional encoding}
The original dot-product attention mechanism does not explicitly take the relative position between different points into account, but it is found that modulating the attention with relative position information between tokens is often beneficial \citep{su2024roformer, oformer2023tmlr}. In this work we propose to use a learnable distance encoding function to directly modulate the dot product between queries and keys:
\begin{equation}
\label{eq:relative distance encoding attention}
    A^{(m)}_{ij} = \sum_{c=1}^{d} \psi^{(m)}_c(e^{(m)}_{ij}) 
    Q^{(m)}_{ic} K^{(m)}_{jc} 
\end{equation}
where  $Q^{(m)}, K^{(m)} \in \mathbb{R}^{S_m\times d}$ are query and key matrices of $m$-th axis (Equation \eqref{general axial attention}), $e^{(m)}_{ij}$ is the absolute angular distance along $m$-th axis between point $i$ and $j$, and $\psi^{(m)}_c: \mathbb{R} \mapsto \mathbb{R}$ is a learnable function. To parameterize $\psi^{(m)}_c$, we use a Bessel basis function with learnable coefficients \citep{gasteiger2019directional}, which previously have shown to be very effective for modelling atomic interactions \citep{gasteiger2019directional, gasteiger2021gemnet} (here we omit superscript $(m)$ for better readability):
\begin{equation}
\label{eq:distance encoding}
    \psi_c(e_{ij})= b_c + \sum_{n=1}^{N_{\text{basis}}} W_{nc} \sqrt{\frac{2}{\pi}} \frac{\sin(n e_{ij})}{e_{ij}},
\end{equation}
where $e_{ij} \in [0, \pi]$ (in practice we add a small number to the left endpoint to avoid overflow), in which the longitude difference $\theta_{ij} = |\theta_i - \theta_j|$ is wrapped into  $[0, \pi]$: $e_{ij}=\min (\theta_{ij}, 2\pi - \theta_{ij})$, and $W_{nc}, b_c$ are a learnable scalar coefficient and offset. For all the experiments, we set $N_{\text{basis}}=64$ along the longitude direction and $N_{\text{basis}}=32$ along the latitude direction. After relative distance encoding, we propose to apply a element-wise non-linearity function- leakyReLU (leaky rectified linear unit) to the attention matrix in Equation \eqref{eq:relative distance encoding attention}, which suppresses the magnitude of the negative part in the matrix and in practice we find this slightly improves the performance. This also coincides with the observation in \citet{performer2020iclr}, where they use ReLU to replace Softmax in the attention kernel and observe performance improvement. 

In addition to the relative distance encoding, we also apply learnable absolute positional encoding to the feature embedding before every Transformer block. The learnable positional encoding is computed by first wrapping the longitude/latitude coordinates with real-valued spherical harmonics and then feeding them to a two-layer MLP. The spherical harmonics can be viewed as an extension of the popular Fourier positional encoding \citep{attention2017nips} on flat Euclidean space to the spherical domain $\mathbb{S}^2$, which generally improves the performance of coordinate-based networks on spheres \citep{russwurm2024geographic, elhag2024manifold}. Example visualization of learned attention kernels, distance and positional encoding are shown in Appendix \ref{appendix more results}.


\vspace{-2mm}
\subsection{Grid projection}

To reduce the computational cost and improve model scalability, the majority of the proposed model (Processor) operates on a latent grid with reduced resolution. For the height dimension (different pressure levels), the Encoder will project 3D variables onto a 2D surface via a height compression module and then in the Decoder the 3D structure of upper-air variables is recovered via a height recovery module. For the spatial dimension (longitudinal and latitudinal), the 2D grid is uniformly downsampled in the Encoder and then re-projected to the original grid via cross-attention in the Decoder. 
\vspace{-2mm}
\paragraph{Height compression/recovery} The height compression is achieved via a 1D height-wise convolution filter, with kernel size equaling the number of pressure levels. To recover the height, we first expand the number of channels of the latent embedding by the number of pressure levels via a linear transformation: $d \mapsto N_{\text{levels}}d$, and then reshape the tensor from $N_{\text{lat}} \times N_{\text{lon}} \times  (N_{\text{levels}}d)$ to $N_{\text{lat}} \times N_{\text{lon}} \times  N_{\text{levels}} \times d$. After recovering the 3D structure, we apply height-wise learnable linear modulation \citep{perez2018film} to the features: $\Tilde{Z}_{\cdots lc} = \alpha_{lc} Z_{\cdots lc} + \beta_{lc}, \quad l \in \{1, 2, \hdots, N_{\text{levels}}\}, c \in \{1, 2, \hdots, d\}$, where $\Tilde{Z}, Z \in \mathbb{R}^{N_{\text{lat}} \times N_{\text{lon}} \times  N_{\text{levels}} \times d}$ are 3D latent embeddings, and $\alpha_{lc}, \beta_{lc}$ are a learnable scalar scaling and offset.
\paragraph{Spatial downsample/upsample} We use patchification to downsample the spatial resolution, which is a widely adopted technique in the vision Transformer \citep{vit2021iclr} to downsample the data resolution and is also used in prior weather forecasting Transformer models. The patchification amounts to applying non-overlapping convolution to the input, with kernel size and stride equal to the patch size $p$ (single-sided zero padding is used when the resolution is not divisible by $p$). This process downsamples the grid resolution from $N_{\text{lat}} \times N_{\text{lon}}$ to $(N_{\text{lat}}/p) \times (N_{\text{lon}}/p)$.

To recover the original resolution, we use factorized cross-attention (Figure \ref{fig:main-arch}(b)) to project the low-resolution latent grid back to the original resolution, which is different from previous works that first apply linear transformation to the channel dimension and then reshape it to "un-patchify" the features. The proposed cross-attention module learns a continuous representation \citep{mildenhall2020nerf} for the target variables and allows querying at arbitrary locations instead of a fixed resolution grid, which can be useful for downstream applications such as data downscaling and regional forecasting. The longitudinal and latitudinal basis functions are paramterized by SIREN \citep{siren2020nips} - a MLP with a sine activation function, which works well in approximating and representing continuous signals, especially their high-frequency components.
\vspace{-2mm}

\subsection{Model overview}
\paragraph{Architecture}
 We use an Encoder-Processor-Decoder (EPD) model architecture \citep{sanchez2020learning, graphcast2023science} (Figure \ref{fig:main-scheme}). The model $F_\theta(\cdot)$ predicts the change of the variable $\Delta X$ over a fixed time interval $\Delta t$ given $X_t$, i.e. $\hat{X}_{t+\Delta t} = f(X_t)=X_t+F_{\theta}(X_t)$. Long-term forecasting is generated by an auto-regressive rollout for multiple steps: 
 \begin{equation}
     \hat{X}_{t+N_T\Delta t}=\underbrace{f \circ f \circ \hdots \circ f}_{\times N_T}(X_t).
 \end{equation}
 
The Encoder takes the surface variables and upper-air variables as input and fuses them together. The 3D upper-air variables are first projected onto a 2D surface via the height compression operation and then concatenated with the surface variable embedding. After concatenation, the surface and upper-air variable embeddings are fused together via a two-layer MLP and then downsampled to the reduced-resolution grid via patchfication.

The Processor is a stack of Transformer blocks (we use 6 blocks across all the experiments) operating on the latent grid,  where each block comprises of a two-layer MLP (also known as feed forward network, FFN) followed by the factorized attention layer. We place FFN in front of attention layer contrary to many existing Transformer architectures that place FFN after the attetion layer. The main reason is that we aim to fuse positional encoding with features via non-linear transformation before doing the attention, which we found performs slightly better. We use residual connections \citep{resnet2016cvpr} between different Transformer blocks and inside the attention layer (Figure \ref{fig:main-arch} (a)). Layer-normalization \citep{ba2016layer} is applied after the residual connection between each Transformer block (which is known as a post-norm scheme in Transformer literatures). We add layer-normalization after projecting features into queries and keys.

The Decoder takes the latent embedding output from the processor (represented on a reduced resolution grid) as input and decodes it into the 2D surface and 3D upper-air variables. The latent embedding is first upsampled via cross-attention, and then passed into two separate MLPs, where the first one is used to decode the embedding to 2D surface variables and the second is used to project and decode the 2D embedding to 3D upper-air variables.

 \paragraph{Training}

We train the model by employing a curriculum that first minimizes the single-step prediction loss and then multi-step prediction loss, similar to prior works \citep{graphcast2023science, chen2023fengwu, chen2023fuxi, nguyen2023stormer, kurth2023fourcastnet}. In the first stage, we train the model for 160k gradient steps, where in the first 50k steps the loss is computed on a single step of prediction-target pairs $(\hat{X}_{t+\Delta t}, X_{t+\Delta t})$, and for the remainng 110k steps the loss is computed based on a two-step rollout of predictions $([\hat{X}_{t+\Delta t}, \hat{X}_{t+2\Delta t} ], [X_{t+\Delta t}, X_{t+2\Delta t}])$. During the second stage, we train the model for 50k gradient steps and increase the model training rollout steps from $4$ to $20$ with a curriculum. The training loss function is a latitude and variable weighted $L1$ loss. The motivation for choosing $L1$ loss over $L2$ loss is to reduce the blurriness of the model's prediction. However, it can make long-term predictions to have higher MSE which we mitigate with multi-step loss during training. The latitude weight and variable weight are computed similar to \citet{graphcast2023science}. The training of the final model on the $1.5^{\circ}$ resolution grid is carried out with 4 A6000 GPUs using Distributed Data Parallel and float32 precision, which takes roughly a week. It takes around 6 seconds to generate the prediction up to a week (28 steps) on a single A6000 using float32 precision, with model wrapped by \texttt{torch.compile}. We provide a brief summary of the model's major hyperparameters in Table \ref{tab: main hyperparameter}, and the model has around 200 M trainable parameters in total. Further details about model training are deferred to the Appendix \ref{appendix model training details}.

The proposed model is developed and validated based on the ERA5 reanalysis data \citep{ERA52020} from the European Centre for Medium-Range Weather Forecasts (ECMWF), where we use the preprocessed version from WeatherBench2 \citep{rasp2023weatherbench2}. Similar to GraphCast \citep{graphcast2023science}, during the model development stage we train the model on $1979-2015$ and validate on $2016$. The final model (on the $1.5^{\circ}$ grid) is trained on $1979-2019$ with data from four additional recent years added to the training dataset. 

\begin{table}[H]
\centering
\begin{tblr}{
  cells = {c},
  hline{1} = {-}{0.08em},
  hline{2} = {-}{0.05em},
  hline{10} = {-}{0.08em},
}
\textbf{Hyperparameter}             & \textbf{Value} \\
Base hidden dimension          & 256            \\
Processor hidden dimension     & 768           \\
Number of Transformer blocks   & 6              \\
Self-attention heads           & 16             \\
Self-attention head dimension  & 128            \\
Cross-attention heads          & 32             \\
Cross-attention head dimension & 128            \\
Latent grid downsample ratio (patch size) & 2$\times$2
\end{tblr}
\vspace{-2mm}
\caption{\label{tab: main hyperparameter} Major hyperparameter choices of the model. \textit{Base hidden dimension} denotes the width of MLPs and number of channels of the latent embedding in Encoder and Decoder. In the patchification layer, the latent embedding's channel size is projected to \textit{Processor hidden dimension} and re-projected back in the Decoder's upsampling (cross-attention) layer. For the FFN inside each Transformer block, the expansion ratio is set to $6$, i.e. the hidden units number is $6$ times the input channel number. Gaussian Linear Error Units (GELUs)\citep{hendrycks2023gaussian} is used as the non-linear activation inside every Transformer block.}
\end{table}
\vspace{-2mm}

\section{Experiments and Results}
\subsection{Evaluation setting}
\begin{table}[H]
\centering
\begin{tblr}{
  cells = {c},
  hline{1} = {-}{0.08em},
  hline{2} = {-}{0.05em},
  hline{8} = {-}{0.08em},
}
\textbf{Constants}                             & \textbf{Surface variables}          & \textbf{Upper-air variables } \\
Land sea mask                         & $2$m temperature  ($t2$m)           & Geopotential  ($zX$)      \\
Soil type                             & $10$m $u$ of wind      ($u10$m)        & Temperature ($tX$)        \\
Angle of sub-gridscale orography      & $10$m $v$ of wind  ($v10$m)            & $u$ of wind ($uX$)          \\
Anisotropy of sub-gridscale orography & Mean sea level pressure  ($\text{MSLP}$)  
& $v$ of wind ($vX$)          \\
                                      & Total precipitation 6hr ($\text{TP}6$) & Specific humidity ($qX$)  \\
                                      & Total precipitation 24hr ($\text{TP}24$)&                     
\end{tblr}
\caption{Weather variables used by the model. Surface and upper-air variables are time-dependent variables predicted by the model given an initial condition. $13$ pressure levels of upper-air variables are considered: $50, 100, 150, 200, 250, 300, 400, 500, 600, 700, 850, 925, 1000~\text{hPa}$. $X$ denotes the pressure level in the abbreviation of upper-air variables, e.g. $z500$ denotes geopotential at $500$hPa. \label{tab: variable summary}}
\end{table}

Variables and their corresponding abbreviations are listed in Table \ref{tab: variable summary}. The time resolution is $\Delta t=6\text{hr}$. The major deterministic metrics are root mean squared error (RMSE), anomaly correlated coefficient (ACC), and bias (see Appendix \ref{appendix metric definition} for definitions). All the metrics are weighted by the corresponding cell area based on latitude \citep{graphcast2023science, rasp2023weatherbench2}: 
\begin{equation}
\label{eq:latitude weight}
   w(i)=\frac{(\sin \varphi^u_i-\sin \varphi^l_i)}{{ \frac{1}{N_{\text{lat}}}}\sum_{j=1}^{N_{\text{lat}}}(\sin \varphi^u_j-\sin \varphi^l_j)} ,
\end{equation}
where $\varphi_u, \varphi_l$ are the corresponding upper/lower latitudinal boundaries for each cell. We exclude the evaluation of total precipitation, as prior works \citep{graphcast2023science, precipitation2022} have suggested the bias of ERA5 precipitation data and its skewed distribution makes deterministic metric like RMSE less meaningful. For MLWP, all the metrics are computed by comparing model's prediction against ERA5 data. For IFS HRES and IFS ENS (ensemble mean), their evaluation metrics are computed by comparing model's prediction against IFS analysis data following WeatherBench 2.

The major model is trained on a $1.5^{\circ}$ resolution equi-angular grid ($240\times 121$) and evaluated on the same grid following the the deterministic evaluation protocol in WeatherBench 2 \citep{rasp2023weatherbench2}.  The initial time for each prediction rollout is 00:00/12:00 UTC of every day in year 2020. To compare our model to the open-sourced weather-forecast Transformer - ClimaX \citep{climax2023icml}, a ViT \citep{vit2021iclr} based model that is customized for climate related tasks with a competitive accuracy on coarse resolution. We also train our model on a $64\times32$ ($5.625^\circ$) grid following the splitting and testing strategy in \citet{climax2023icml}. Other MLWP baseline models considered are Pangu-Weather (Pangu) \citep{pangu2023nature}, FuXi \citep{chen2023fuxi} and GraphCast \citep{graphcast2023science}, which represent the state-of-the-art purely data-driven MLWP. All three models are trained on a $0.25$ resolution grid with evaluation results projected to a $1.5^\circ$ grid \citep{rasp2023weatherbench2}. Pangu trains different lead time models ($1h, 3h, 6h, 24h$) and cascade them during inference to generate a final prediction. FuXi scales up the number of Swin Transformer \citep{swin2021cvpr} blocks compared to Pangu (48 vs 16). Instead of cascading models with different lead times, FuXi uses a fixed $6h$  lead time but fine-tunes separate models for different time scales and cascade them during prediction. GraphCast is built upon a multi-resolution graph on the spherical geometry and uses a model with lead time $6h$ to autogressively generate predictions.

\subsection{Discussion}

The quantitative results for different models' prediction accuracies are presented in Table \ref{tab: rmse results 1} and \ref{tab: rmse results 2}. For pressure variables ($z500$, MSLP), we observe that CaFA consistently outperforms other MLWP baseline within the range of 5 days. For other target variables, CaFA falls behind GraphCast and FuXi for 1 day prediction but gradually catches up. We hypothesize the primary reason is that compared to these models, CaFA is trained on lower resolution data ($1.5^\circ$ vs $0.25^\circ$) and therefore does not have access to finer scale physics which can be important for features like the wind velocity field. The influence of finer-scale physics will gradually diminish as time evolves. This is due to the chaotic nature of weather dynamics and deterministic ML models will tend to approximate the statistical mean which usually dampens the high frequency components \citep{bonavita2023limitations}. In general, CaFA is on par with other MLWP baselines for prediction from 3 days up to a week and performs particularly well in variables that features a decaying spectrum such as geopotential. It is also a more lightweight Transformer-based model compared to FuXi, which uses 48 Transformer blocks with hidden dimension $1536$ (each block comprises an attention layer and a FFN) in total whereas CaFA uses 6 Transformer blocks plus 1 cross-attention layer with hidden dimension $768$. 

\begin{table}[H]
\centering
\fontsize{9.5pt}{10pt}\selectfont
\setlength{\tabcolsep}{2.5pt}
\begin{NiceTabular}{lcccccccccccccccc}
\toprule
\textbf{Variables} & \multicolumn{4}{c}{$\textbf{z500}$ {\small[kg$^2/$m$^2$]}} & \multicolumn{4}{c}{$\textbf{t850}$ {\small[K]}} & \multicolumn{4}{c}{$\textbf{t2}$\textbf{m} {\small[K]}} & \multicolumn{4}{c}{$\textbf{MSLP}$ {\small[Pa]}} \\ 
\cmidrule(lr){2-5}
\cmidrule(lr){6-9}
\cmidrule(lr){10-13}
\cmidrule(lr){14-17}

\textit{Lead hours} & \textit{24} & \textit{72} & \textit{120} & \textit{168} & \textit{24} & \textit{72} & \textit{120} & \textit{168} & \textit{24} & \textit{72} & \textit{120} & \textit{168} & \textit{24} & \textit{72} & \textit{120} & \textit{168} \\ 
\midrule
IFS HRES & \rmsepressureone{42} & \rmsepressurethree{135} & \rmsepressurefive{304} & \rmsepressureseven{521} & \rmsetemperatureone{0.62} & \rmsetemperaturethree{1.15} & \rmsetemperaturefive{1.79} & \rmsetemperatureseven{2.59} & \rmsetsurfaceone{0.51} & \rmsetsurfacethree{0.87} & \rmsetsurfacefive{1.33} & \rmsetsurfaceseven{1.88} & 

\rmsepressuresurfaceone{60} & \rmsepressuresurfacethree{149} & \rmsepressuresurfacefive{309} & \rmsepressuresurfaceseven{506} \\

IFS ENS & \rmsepressureone{42} & \rmsepressurethree{132} & \rmsepressurefive{277} & \rmsepressureseven{439} & \rmsetemperatureone{0.63} & \rmsetemperaturethree{1.09} & \rmsetemperaturefive{1.59} & \rmsetemperatureseven{2.14} & \rmsetsurfaceone{0.56} & \rmsetsurfacethree{0.86} & \rmsetsurfacefive{1.20} & \rmsetsurfaceseven{1.58} & 

\rmsepressuresurfaceone{62} & \rmsepressuresurfacethree{146} & \rmsepressuresurfacefive{280} & \rmsepressuresurfaceseven{423} \\ 
\midrule
CaFA(Ours) & 
\rmsepressureone{39} & \rmsepressurethree{123} & \rmsepressurefive{269} & \rmsepressureseven{449} & 

\rmsetemperatureone{0.60} & \rmsetemperaturethree{0.99} & \rmsetemperaturefive{1.56} & \rmsetemperatureseven{2.28} &

\rmsetsurfaceone{0.55} & \rmsetsurfacethree{0.84} & \rmsetsurfacefive{1.24} & \rmsetsurfaceseven{1.76} &

\rmsepressuresurfaceone{48} & \rmsepressuresurfacethree{130} & \rmsepressuresurfacefive{271} & 
\rmsepressuresurfaceseven{438}\\

Pangu &
\rmsepressureone{44} & \rmsepressurethree{133} & \rmsepressurefive{294} & \rmsepressureseven{501} & 

\rmsetemperatureone{0.61} & \rmsetemperaturethree{1.03} & \rmsetemperaturefive{1.68} & \rmsetemperatureseven{2.47} & 

\rmsetsurfaceone{0.56} & \rmsetsurfacethree{0.91} & \rmsetsurfacefive{1.39} & \rmsetsurfaceseven{1.97} & 

\rmsepressuresurfaceone{55} & \rmsepressuresurfacethree{143} & \rmsepressuresurfacefive{297} & 
\rmsepressuresurfaceseven{486} \\
FuXi &

\rmsepressureone{40} & \rmsepressurethree{125} & \rmsepressurefive{276} & \rmsepressureseven{433} & 

\rmsetemperatureone{0.54} & \rmsetemperaturethree{0.95} & \rmsetemperaturefive{1.56} & \rmsetemperatureseven{2.11} & 

\rmsetsurfaceone{0.53} & \rmsetsurfacethree{0.84} & \rmsetsurfacefive{1.27} & \rmsetsurfaceseven{1.65} & 

\rmsepressuresurfaceone{49} & 
\rmsepressuresurfacethree{132} & \rmsepressuresurfacefive{276} & \rmsepressuresurfaceseven{416} \\

GraphCast &

\rmsepressureone{39} & \rmsepressurethree{124} & \rmsepressurefive{274} & \rmsepressureseven{468} & 

\rmsetemperatureone{0.51} & \rmsetemperaturethree{0.93} & \rmsetemperaturefive{1.53} & \rmsetemperatureseven{2.29} & 

\rmsetsurfaceone{0.51} & \rmsetsurfacethree{0.82} & \rmsetsurfacefive{1.25} & \rmsetsurfaceseven{1.81} & 

\rmsepressuresurfaceone{48} & 
\rmsepressuresurfacethree{132} & \rmsepressuresurfacefive{276} &\rmsepressuresurfaceseven{454} \\ 

\midrule
Climatology &\cellcolor{low!75}820 &\cellcolor{low!75}820 & \cellcolor{low!75}820 &\cellcolor{low!75}820 & \cellcolor{low!75}3.41 & \cellcolor{low!75}3.41 & \cellcolor{low!75}3.41 & \cellcolor{low!75}3.41 & \cellcolor{low!75}2.61 & \cellcolor{low!75}2.61 & \cellcolor{low!75}2.61 & \cellcolor{low!75}2.61 & \cellcolor{low!75}725 & \cellcolor{low!75}725 & \cellcolor{low!75}725 & \cellcolor{low!75}725 \\
\bottomrule
\end{NiceTabular}
\vspace{-1mm}
\caption{Deterministic forecast RMSE on year $2020$'s data for temperature and pressure at different lead times. Darker colors indicate better performance. The evaluation results for baseline models are taken from WeatherBench2 \citep{rasp2023weatherbench2}. \label{tab: rmse results 1}}
\end{table}%
\begin{table}[H]
\centering
\fontsize{9.5pt}{10pt}\selectfont
\setlength{\tabcolsep}{1.6pt}
\begin{NiceTabular}{lcccccccccccccccccccc}
\toprule
\textbf{Variables} &
\multicolumn{4}{c}{$\textbf{q700}$ {\small[g/kg]}} & 
\multicolumn{4}{c}{$\textbf{u850}$ {\small[m/s]}} &
\multicolumn{4}{c}{$\textbf{v850}$ {\small[m/s]}} &
\multicolumn{4}{c}{$\textbf{u10m}$ {\small[m/s]}} &
\multicolumn{4}{c}{$\textbf{v10m}$ {\small[m/s]}} \\ 
\cmidrule(lr){2-5}
\cmidrule(lr){6-9}
\cmidrule(lr){10-13}
\cmidrule(lr){14-17}
\cmidrule(lr){18-21}

\textit{Lead hours} & \textit{24} & \textit{72} & \textit{120} & \textit{168} & \textit{24} & \textit{72} & \textit{120} & \textit{168} & \textit{24} & \textit{72} & \textit{120} & \textit{168} & \textit{24} & \textit{72} & \textit{120} & \textit{168} & 
\textit{24} & \textit{72} & \textit{120} & \textit{168} \\ 
\midrule
IFS HRES & 
\rmsehumidityone{0.56} & \rmsehumiditythree{0.97} & \rmsehumidityfive{1.28} & \rmsehumidityseven{1.54} & 

 \rmseuwindone{1.18} &  \rmseuwindthree{2.32} &  \rmseuwindfive{3.65} & \rmseuwindseven{4.99} &
 
\rmsevwindone{1.20} & \rmsevwindthree{2.34} & \rmsevwindfive{3.69} & \rmsevwindseven{5.07} &

\rmseusurfaceone{0.83} & \rmseusurfacethree{1.61} & \rmseusurfacefive{2.56} & \rmseusurfaceseven{3.50} &

\rmsevsurfaceone{0.87} & \rmsevsurfacethree{1.67} & \rmsevsurfacefive{2.66} & \rmsevsurfaceseven{3.66}
 \\
 IFS ENS & 
\rmsehumidityone{0.51} & \rmsehumiditythree{0.84} & \rmsehumidityfive{1.06} & \rmsehumidityseven{1.22} & 

 \rmseuwindone{1.14} & \rmseuwindthree{2.10} & \rmseuwindfive{3.13} & \rmseuwindseven{4.03} &
 
\rmsevwindone{1.16} & \rmsevwindthree{2.12} & \rmsevwindfive{3.16} & \rmsevwindseven{4.08} &

\rmseusurfaceone{0.81} & \rmseusurfacethree{1.47} & \rmseusurfacefive{2.19} & \rmseusurfaceseven{2.82} &

\rmsevsurfaceone{0.84} & \rmsevsurfacethree{1.51} & \rmsevsurfacefive{2.27} & \rmsevsurfaceseven{2.95}
 \\
\midrule

CaFA(Ours) &
\rmsehumidityone{0.55} & \rmsehumiditythree{0.84} & \rmsehumidityfive{1.07} & \rmsehumidityseven{1.30} & 

 \rmseuwindone{1.11} & \rmseuwindthree{1.98} & 
 \rmseuwindfive{3.09} & \rmseuwindseven{4.24} &
 
 \rmsevwindone{1.12} & \rmsevwindthree{2.00} & \rmsevwindfive{3.13} & \rmsevwindseven{4.30} &
 
\rmseusurfaceone{0.69} & \rmseusurfacethree{1.34} & \rmseusurfacefive{2.16} & \rmseusurfaceseven{2.98} &

\rmsevsurfaceone{0.72} & \rmsevsurfacethree{1.39} & \rmsevsurfacefive{2.24} & \rmsevsurfaceseven{3.12} \\

Pangu &
\rmsehumidityone{0.54} & \rmsehumiditythree{0.88} & \rmsehumidityfive{1.20} & \rmsehumidityseven{1.48} & 

 \rmseuwindone{1.17} & \rmseuwindthree{2.11} &
 \rmseuwindfive{3.38} & \rmseuwindseven{4.70} &
 
\rmsevwindone{1.19} & \rmsevwindthree{2.15} & \rmsevwindfive{3.44} & \rmsevwindseven{4.79} &

\rmseusurfaceone{0.73} & \rmseusurfacethree{1.43} & \rmseusurfacefive{2.36} & \rmseusurfaceseven{3.28} &

\rmsevsurfaceone{0.76} & \rmsevsurfacethree{1.49} & \rmsevsurfacefive{2.45} & \rmsevsurfaceseven{3.44} \\

FuXi &
- & - & - & - & 
 \rmseuwindone{1.03} & \rmseuwindthree{1.96} &  \rmseuwindfive{3.15} & \rmseuwindseven{3.96} &
 
\rmsevwindone{1.05} & \rmsevwindthree{2.00} & \rmsevwindfive{3.21} & \rmsevwindseven{4.03} &

\rmseusurfaceone{0.66} & \rmseusurfacethree{1.33} & \rmseusurfacefive{2.19} & \rmseusurfaceseven{2.74} &

\rmsevsurfaceone{0.69} & \rmsevsurfacethree{1.38}
& \rmsevsurfacefive{2.28} & \rmsevsurfaceseven{2.87} \\

GraphCast &
\rmsehumidityone{0.47} & \rmsehumiditythree{0.80} & \rmsehumidityfive{1.06} & \rmsehumidityseven{1.31} &

 \rmseuwindone{1.00} & \rmseuwindthree{1.94} & 
  \rmseuwindfive{3.12} & \rmseuwindseven{4.36} &
 
\rmsevwindone{1.02} & \rmsevwindthree{1.97} & \rmsevwindfive{3.17} & \rmsevwindseven{4.43} &

\rmseusurfaceone{0.65} & \rmseusurfacethree{1.32} & \rmseusurfacefive{2.18} & \rmseusurfaceseven{3.05} &

\rmsevsurfaceone{0.68} & \rmsevsurfacethree{1.37} &
\rmsevsurfacefive{2.26} & \rmsevsurfaceseven{3.19} \\

\midrule

Climatology &
\cellcolor{low!75}1.61 &\cellcolor{low!75}1.61 &\cellcolor{low!75}1.61 & \cellcolor{low!75}1.61 & 

\cellcolor{low!75}5.66 & \cellcolor{low!75}5.66 & \cellcolor{low!75}5.66 & \cellcolor{low!75}5.66 &

\cellcolor{low!75}5.47 & \cellcolor{low!75}5.47 & \cellcolor{low!75}5.47 & \cellcolor{low!75}5.47 &

\cellcolor{low!75}3.83 & \cellcolor{low!75}3.83 & 
\cellcolor{low!75}3.83 & \cellcolor{low!75}3.83 &

\cellcolor{low!75}3.96 & \cellcolor{low!75}3.96  &\cellcolor{low!75}3.96 & \cellcolor{low!75}3.96  \\

\bottomrule
\end{NiceTabular}
\vspace{-2mm}
\caption{Deterministic forecast RMSE for specific humidity and wind speed components at different lead times and pressure levels on year 2020's data.\label{tab: rmse results 2}}
\end{table}%

\begin{figure}[H]
    \centering
    \vspace{-2mm}
    \includegraphics[width=0.99\textwidth]{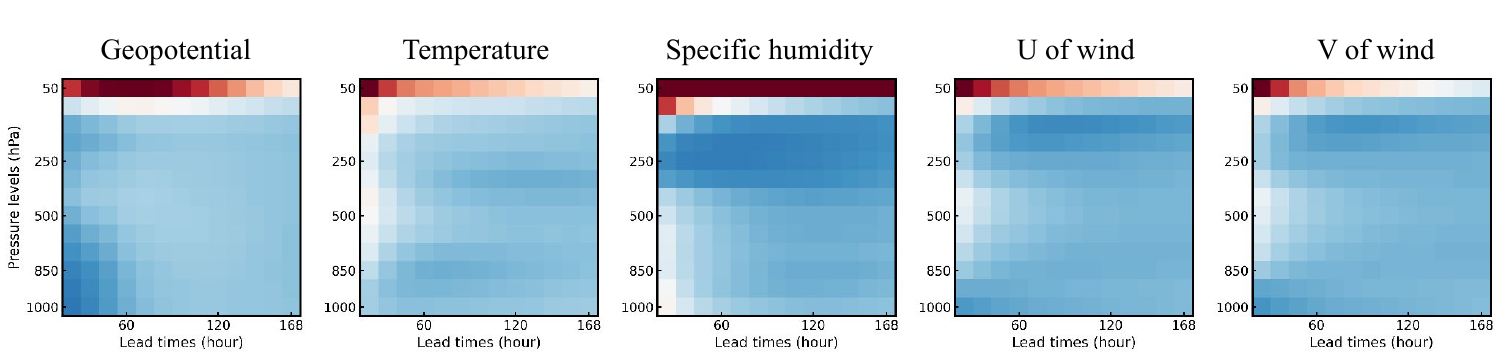}
    \vspace{-2mm}
    \caption{\label{fig:upper air rmse}
Normalized RMSE difference: {\small $(\text{RMSE}_{\text{HRES}} - \text{RMSE}_{\text{CaFA}}) / \text{RMSE}_{\text{HRES}}$}. Blue colors indicate IFS HRES has larger RMSE while red colors indicate CaFA has larger RMSE. Darker colors indicate a larger normalized difference. The plotting style follows GraphCast \citep{graphcast2023science}.}
\end{figure}

\begin{figure}[H]
    \centering
    \vspace{-3mm}
    \includegraphics[width=0.95\textwidth]{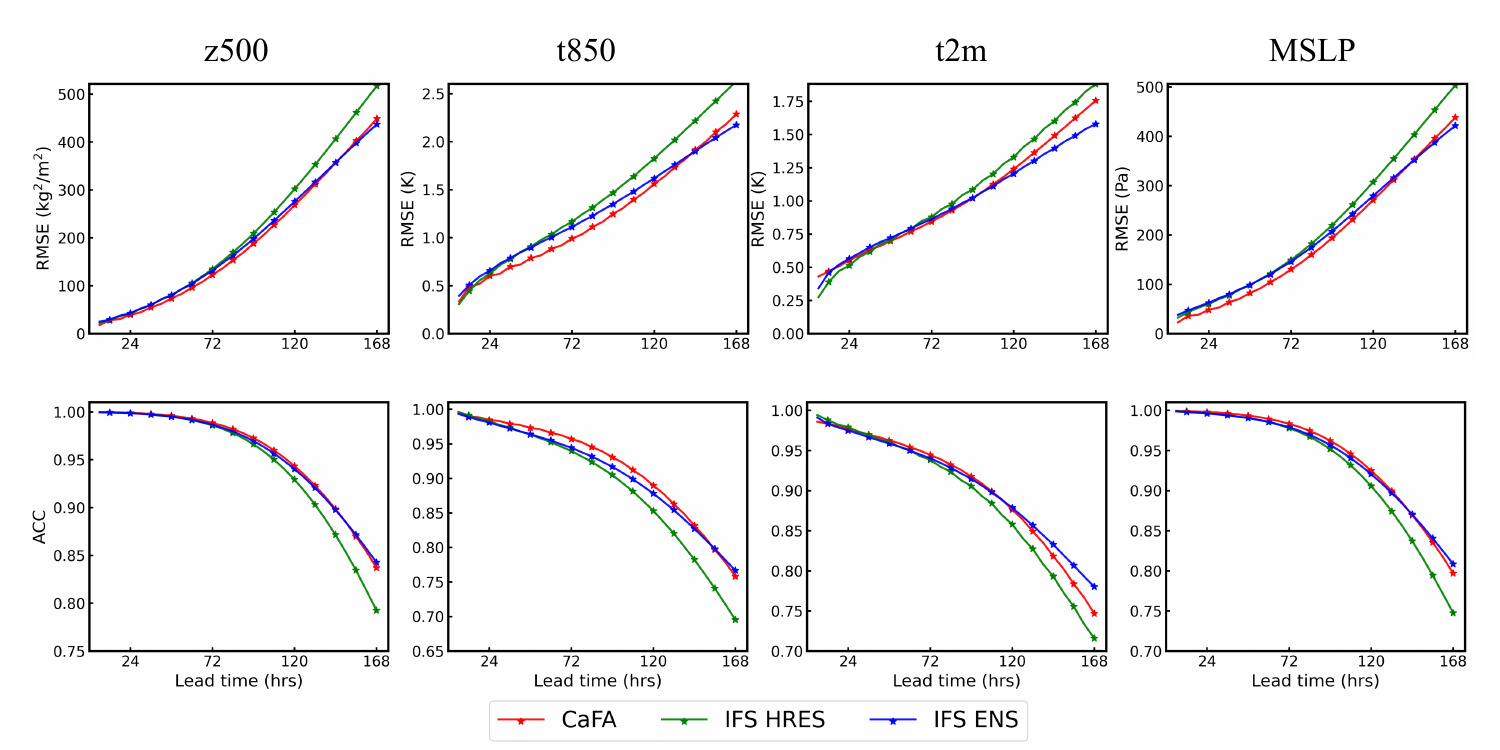}
    \vspace{-2mm}
    \caption{Comparison of CaFA and NWP in temperature and pressure's prediction of year 2020. \label{fig:error trend 1}}
\end{figure}

\begin{figure}[H]
    \centering
     \includegraphics[width=\textwidth]{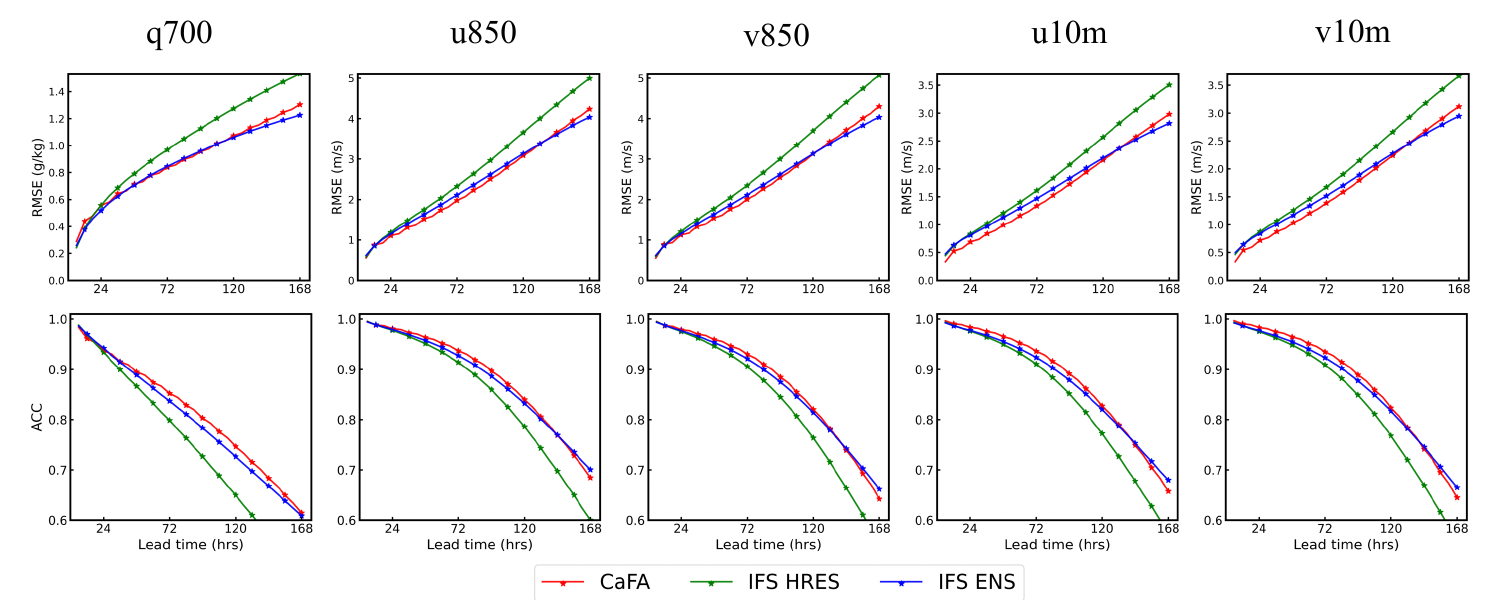}
     \vspace{-2mm}
     \caption{Comparison of CaFA and NWP in specific humidity and wind velocity prediction of year 2020\label{fig:error trend 2}.}
\end{figure}

As shown in Figure \ref{fig:error trend 1} and \ref{fig:error trend 2}, CaFA has lower RMSE and higher ACC than IFS HRES from $1$ to $7$ days on most of the target variables. Comparing to the ensemble mean of IFS forecasts, CaFA's performance is on par or better up to 5 days, but after 5 days the ensemble mean generally has better accuracy. We find that for CaFA, using a larger time step size can improve forecasting accuracy beyond 7 days by trading off accuracy within 7 days range (the comparison is shown in Appendix \ref{fig:6hvs12h}). We also analyze the upper-air variables' prediction of different pressure levels by comparing it to IFS HRES prediction. Figure \ref{fig:upper air rmse} shows that CaFA has better RMSE across most pressure levels except for $50$hPa, where HRES consistently performs better. Such a trend is also observed in GraphCast \citep{graphcast2023science} and one of the reason is that both CaFA and GraphCast are trained on loss that outweighs near-surface (higher pressure) accuracy over smaller pressure levels' accuracy. An example rollout visualization of 3 key variables is shown in Figure \ref{fig:qualitative visualization}, qualitatively the models' prediction agrees well with reference ERA5 analysis data in terms of the general pattern. On the wind velocity field, it is observed that the high-frequency details are blurrier as lead time increases. Visualizations of other variables are shown in Appendix \ref{appendix more results}.

We also compare our model to ClimaX \citep{climax2023icml} \footnote{ClimaX's official code for model and its training is publicly available (with MIT license) at: \url{https://github.com/microsoft/ClimaX/tree/main}.} to better understand the influence of different attention mechanisms. To factor out the influence of discretization (in space and time) and feature selection, we slightly modify the original ViT-based architecture in ClimaX to train an autoregressive model (the resulting model is called ClimaX-AR) with lead time equal to $6$ hours using the weather variables specified in Table \ref{tab: variable summary} (which uses more atmospheric variables than original ClimaX). This ensures an exact same training pipeline between CaFA and ClimaX-AR. We observe Climax-AR performs notably better than the direct prediction version used in original ClimaX, particularly at shorter lead times. As shown in Table \ref{tab: CaFA vs climax}, equipped with factorized attention, CaFA performs consistently better on four key variables for lead time 1-7 days, using significantly smaller amount of FLOPs. We also provide a computational benchmark comparing factorized attention and standard attention on different resolution grid in the Appendix \ref{fig:computational benchmark}, which shows that factorized attention greatly reduces the computational cost associated with higher resolution grid.  This highlights that the potential of proposed factorized attention in improving the efficiency/accuracy balance of Transformer-based weather models.

\begin{table}[H]
\vspace{-2mm}
\centering
\fontsize{10.5pt}{11pt}\selectfont
\setlength{\tabcolsep}{2.1pt}
\begin{NiceTabular}{lccccccccccccccccc}
\toprule
\textbf{Variables} & \multicolumn{4}{c}{$\textbf{z500}$ {\small[kg$^2/$m$^2$]}} & \multicolumn{4}{c}{$\textbf{t850}$ {\small[K]}} & \multicolumn{4}{c}{$\textbf{t2}$\textbf{m} {\small[K]}} & \multicolumn{4}{c}{$\textbf{u10}$\textbf{m} {\small[m/s]}} & \multirow{2}{*} {TFLOPs}\\ 
\cmidrule(lr){2-5}
\cmidrule(lr){6-9}
\cmidrule(lr){10-13}
\cmidrule(lr){14-17}

\textit{Lead hours} & \textit{24} & \textit{72} & \textit{120} & \textit{168} & \textit{24} & \textit{72} & \textit{120} & \textit{168} & \textit{24} & \textit{72} & \textit{120} & \textit{168} & \textit{24} & \textit{72} & \textit{120} & \textit{168} & \\ 
\midrule

CaFA*(Ours)&
\bcpressureone{51} & \bcpressurethree{170} & \bcpressurefive{348} & \bcpressureseven{544} & 

 \bctempone{0.59} & \bctempthree{1.04} & 
 \bctempfive{1.71} &  \bctempseven{2.47} &
 
\bctempsurfone{0.49} & \bctempsurfthree{0.82} & \bctempsurffive{1.28} & \bctempsurfseven{1.79} & 

 \bcusurfone{0.58} & \bcusurfthree{1.28} & 
 \bcusurffive{2.15} & \bcusurfseven{2.97} & \bcflops{0.61}
 \\ 

ClimaX-AR &
 \bcpressureone{71} & \bcpressurethree{215} & \bcpressurefive{404} & \bcpressureseven{585} & 
 
  \bctempone{0.73} & \bctempthree{1.25} & 
  \bctempfive{1.93} & \bctempseven{2.62} & 
  
\bctempsurfone{0.56} & \bctempsurfthree{0.95} & \bctempsurffive{1.42} & \bctempsurfseven{1.90} &

 \bcusurfone{0.69} & \bcusurfthree{1.50} & \bcusurffive{2.37}  & \bcusurfseven{3.08} 
 & \bcflops{2.22}
 \\ 

ClimaX* &
 \bcpressureone{96} & \bcpressurethree{244} & \bcpressurefive{440} & \bcpressureseven{599} &
 
  \bctempone{1.11} & \bctempthree{1.59} & 
  \bctempfive{2.23} &\bctempseven{2.77}  &
  
\bctempsurfone{1.10} & \bctempsurfthree{1.43} & 
\bctempsurffive{1.83} & \bctempsurfseven{2.18} & 

 \bcusurfone{1.41} & \bcusurfthree{2.18} & \bcusurffive{2.94} & \bcusurfseven{3.43} &
- 
 \\ 
\bottomrule
\end{NiceTabular}
\vspace{-1mm}
\caption{\label{tab: CaFA vs climax} Comparison of accuracy and computational cost against ClimaX \citep{climax2023icml} on a $5.625^{\circ}$ resolution grid. The data splitting follows the strategy used in ClimaX, where the test years are $2017, 2018$. ClimaX* is the direct prediction model reported in the original paper (Table 10) and uses fewer atmospheric variables than our framework. T (tera) FLOPs (number of floating point operations) of models' forward propagation is measured using the DeepSpeed profiling utility \citep{deepspeed2020} with PyTorch 2.1 and a batch size of 8. CaFA* is a smaller version of the major model, with a processor hidden dimension of $512$ ($2/3$ of the number specified in Table \ref{tab: main hyperparameter}) and the total trainable amount of parameters therefore roughly matches ClimaX ($\sim$ 100M). The number of attention heads is $16$ for both ClimaX and CaFA, while the hidden dimension of each attention head is $128$ for CaFA and $64$ for ClimaX.}
\end{table}%
\begin{figure}[H]
    \centering
    \vspace{-2mm}
    \begin{subfigure}{\textwidth}
    \includegraphics[width=0.99\textwidth]{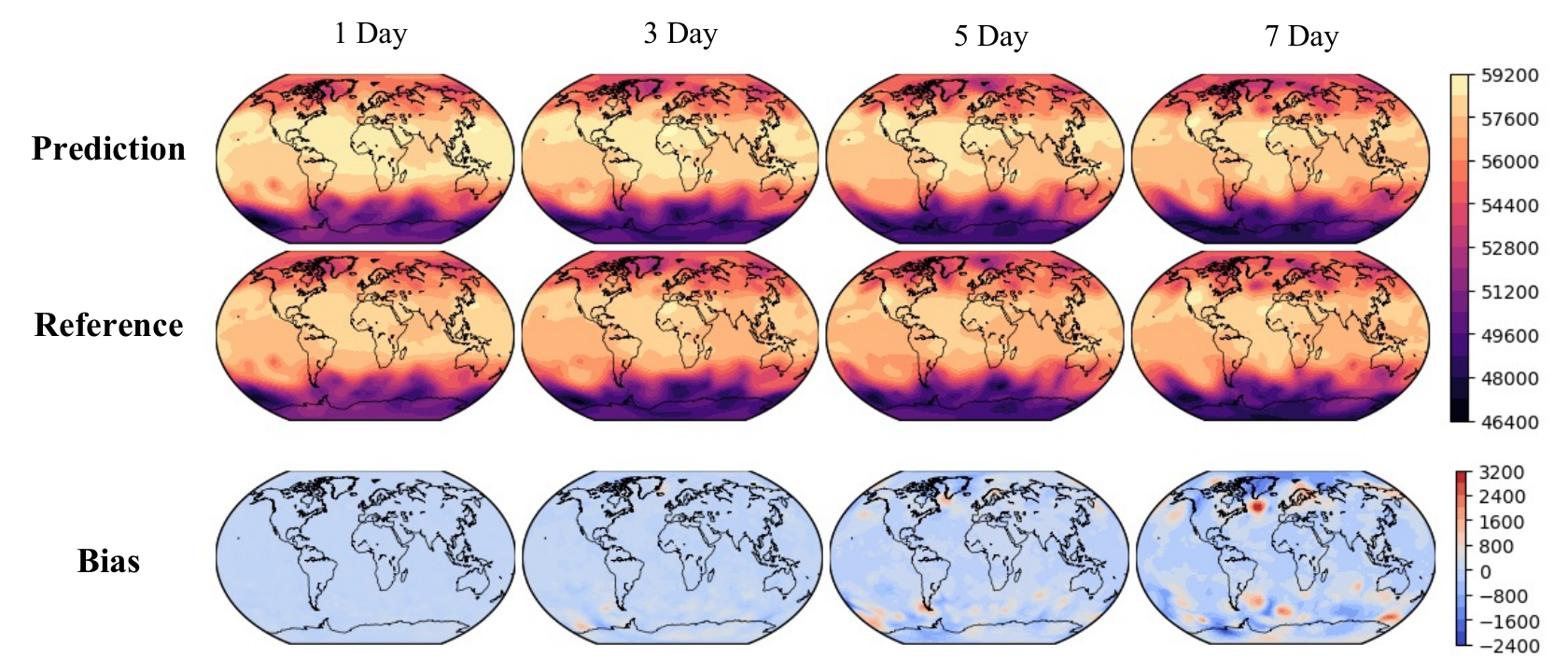}
    \vspace{-2mm}
    \caption{$z500$ example rollout visualization}
    \vspace{-5mm}
    \end{subfigure}  
\end{figure}%
\begin{figure}[H]\ContinuedFloat 
    \begin{subfigure}{\textwidth}
    \includegraphics[width=0.99\textwidth]{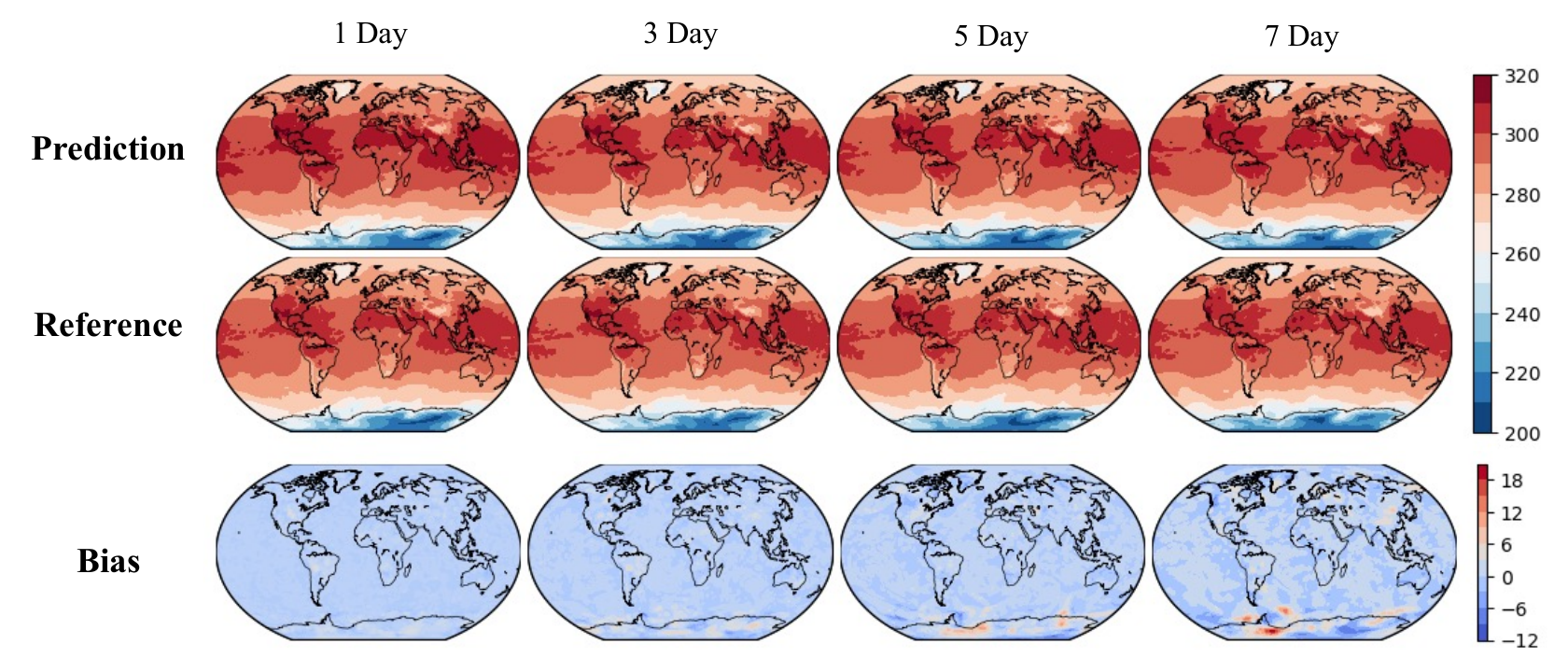}
    \vspace{-2mm}
    \caption{$t2$m example rollout visualization}
    \end{subfigure}
    \vspace{-2mm}
\end{figure}%
\begin{figure}[H]\ContinuedFloat 
    \begin{subfigure}{\textwidth}
    \vspace{-2mm}
    \includegraphics[width=0.99\textwidth]{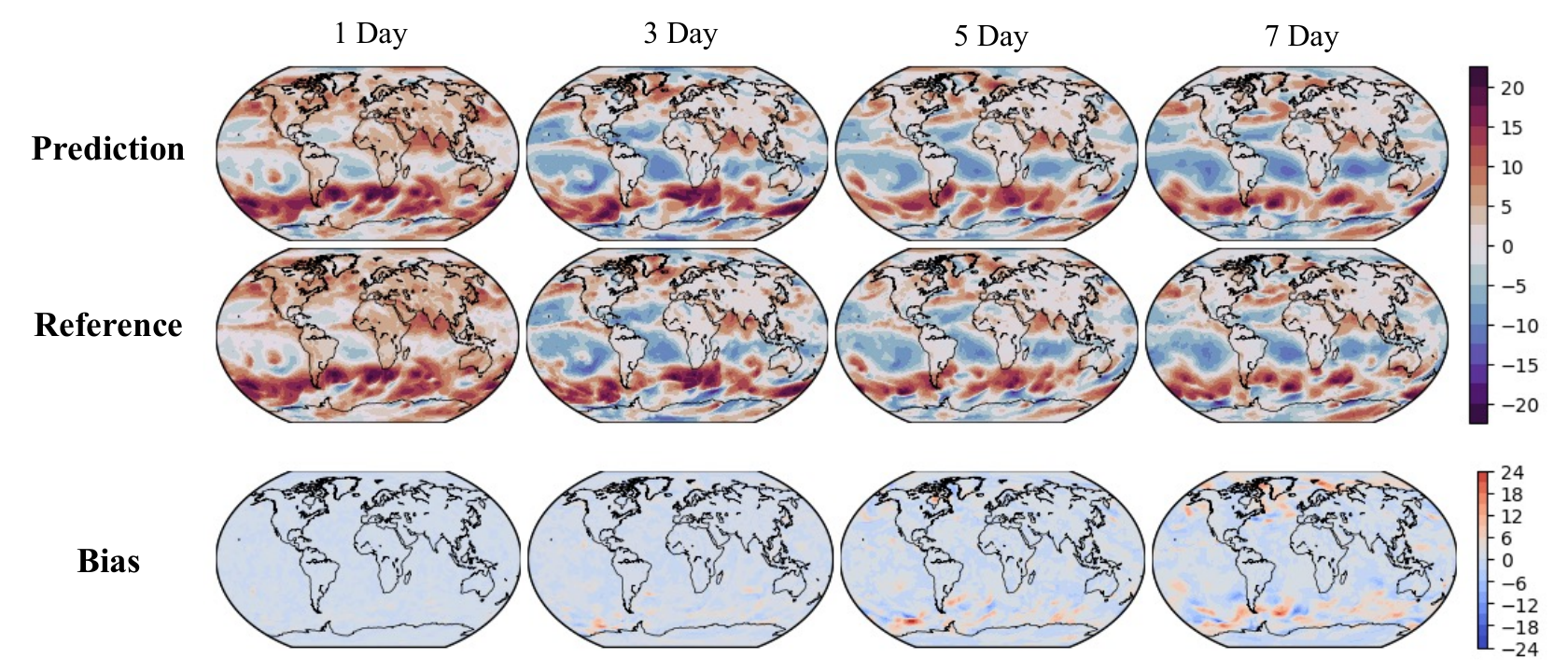}
    \vspace{-1mm}
    \caption{$u10$m example rollout visualization}
    \end{subfigure}
    \vspace{-2mm}
    \caption{\label{fig:qualitative visualization} Example rollout visualizations of the model's prediction versus reference ERA5 reanalysis data at different lead times. The initialization time is 00:00 UTC on August 11, 2020. }
\end{figure}

\section{Conclusion}

In this work we propose an end-to-end machine learning weather prediction model based on factorized attention on a sphere. Our work presents a computationally efficient methodology for computing a data-controlled attention kernel that is capable of capturing long-range dependencies in the domain. The factorized axial kernels only have a single-dimensional domain and thus their computation is very efficient. The proposed attention mechanism respects the underlying periodic boundary along longitudinal direction and responds smoothly with respect to the angular distance between spatial points. Our work signifies an advancement in enhancing the efficiency and accuracy of Transformer-based weather prediction models.

Due to the limited computational budget, the finest resolution we have studied in this work is a spatial resolution of $1.5^\circ$ and temporal resolution of $6$ hours. Scaling the model to higher resolutions in both space and time can potentially allow the model to capture finer scale physics and provide better predictions. In parallel, it is interesting to study the scaling of model itself to test how much accuracy improvement can be achieved by using more compute. In addition, we have only explored using factorized attention to model spatial interactions, which works particularly well for physical processes that have a decaying spectrum such as geopotential. To further improve the learning capability of higher-frequency components, it is also an interesting direction to combine the proposed model with other localized operators such as message passing around local neighborhoods.

In the current study, we only focus on investigating the short to medium range deterministic forecast capability of the proposed model. However, due to the chaotic nature of weather dynamics, deterministic and purely data-driven models tend to generate blurry predictions in the long run as a result of increasing uncertainty. The IFS ensemble prediction is generally observed to have superior accuracy in longer range forecasting compared to deterministic MLWP. We anticipate that the integration of probabilistic frameworks and ensemble prediction that accounts for the inherent uncertainty will be an important next step. It is also worth pointing out the current model is a purely data-driven approach and it is not guaranteed to satisfy physics constraints. Hence, incorporating more physics priors and symmetries into the model can also be a meaningful future direction.


\bibliographystyle{abbrvnat}
\bibliography{_bibliography/attention_and_transformer, _bibliography/neural_operator, _bibliography/misc, _bibliography/ml_weather, _bibliography/classical_weather}

\clearpage
\appendix
\rule[0pt]{\columnwidth}{1pt}
\begin{center}
    \huge
    \emph{Supplementary Material}
\end{center}
\rule[0pt]{\columnwidth}{1.5pt}
\doparttoc
\tableofcontents
\newpage
\section{Model implementation}
\subsection{Training details}
\label{appendix model training details}

We specify the detailed hyperparamters for the model training in Table \ref{tab:train hyperparameter stage 1} and \ref{tab:train hyperparameter stage 2}. We implement the model and the experiment using PyTorch \citep{paszke2019pytorch}. We use AdamW optimizer \citep{adam15, adamw2018} and the majority of the hyperparameter setting follows ClimaX \citep{climax2023icml}. Regularization techniques including dropout \citep{dropout2014} and droppath \citep{larsson2016fractalnet} are not used in the proposed model (which is used in ClimaX baseline). For testing, we evaluate the final model checkpoint and no early stopping is applied. During second-stage training, gradient checkpointing is used to reduce the memory cost of backpropagating through time. In addition, only the most recent 20 years' (up to the validation years) training data is used for second stage training.

\begin{table}[H]
\centering
\begin{tblr}{
  cells = {c},
  hline{1} = {-}{0.08em},
  hline{2} = {-}{0.05em},
  hline{12} = {-}{0.08em},
}
\textbf{Training Hyperparameter}             & \textbf{Value} \\
Optimizer          & AdamW            \\
Learning rate     & $1e-8 \mapsto 3e-4 \mapsto 1e-7$          \\
Batch size   & 8              \\
Gradient steps           & 160k            \\
Learning rate scheduler  & Cosine Annealing with \\ & Linear Warmup (for 1600 steps)            \\
Weight decay          & 1e-6             \\
Rollout step curriculum & $[1, 2]$          \\
Curriculum milestone & $[0, 50\text{k}] $           \\
EMA & None
\end{tblr}
\caption{\label{tab:train hyperparameter stage 1} Training hyperparameter for stage 1. On $64\times32$ grid we adjust the batch size to $16$ and peak learning rate to $5e-4$. The curriculum milestone determines how many steps to rollout after specific milestone. In the above setting, after 50k gradient steps, the model will rollout for 2 steps.}
\end{table}

\begin{table}[H]
\centering
\begin{tblr}{
  cells = {c},
  hline{1} = {-}{0.08em},
  hline{2} = {-}{0.05em},
  hline{12} = {-}{0.08em},
}
\textbf{Training Hyperparameter}             & \textbf{Value} \\
Optimizer          & AdamW            \\
Learning rate     & $1e-8 \mapsto 3e-7 \mapsto 1e-7$          \\
Batch size   & 8              \\
Gradient steps           & 50k            \\
Learning rate scheduler  & Cosine Annealing with \\ & Linear Warmup (for 500 steps)            \\
Weight decay          & 1e-6             \\
Rollout step curriculum & $[4, 8, 12, 16, 20]$          \\
Curriculum milestone & $[0, 10\text{k}, 20\text{k}, 30\text{k}, 40\text{k}] $           \\
EMA & Deacy=0.999
\end{tblr}
\caption{\label{tab:train hyperparameter stage 2} Training hyperparameter for stage 2. We save the Exponential Moving Average (EMA) of model weights during the second stage.}
\end{table}
\newpage

The training loss is a latitude-weighted and variable-weighted $L1$ loss:
\begin{equation}
    \label{eq:training loss}
    \text{Loss}(\hat{X}, X) =
     \frac{1}{N_{\text{lat}}N_\text{lon}N_\text{t}}
    \sum_{c=1}^{N_c}\lambda_c
    \sum_{l=1}^{N_\text{level}}\lambda_l
    \sum_{t=1}^{N_{\text{t}}}
    \sum_{i=1}^{N_{\text{lat}}}
    \sum_{j=1}^{N_{\text{lon}}}
    w(i)||\hat{X}_{tijl}^c-X_{tijl}^c||
    ,
\end{equation}
where $w(i)$ is the latitude-based weight (see Equation \ref{eq:latitude weight}),  $\lambda_l$ is the level-based weight, $\lambda_c$ is the variable-based weight,  $\hat{X}$ is model's prediction and $X$ is reference data, $N_t$ denotes total number of rollout timesteps, $N_{\text{lat}}$ denotes number of latitude grid points, $N_{\text{lon}}$ denotes number of longitude grid points. The level-based weight is a simple linear interpolation between $0.05$ and $0.065$: $\lambda_l=\text{Index}(l)/12 \times 0.015 + 0.050$, where the index of pressure level $l$ ranges from 0 to 12 corresponds to 50hPa to 1000 hPa.  We list the specification of the variable-based weight below (see Table \ref{tab: variable summary} for the meaning of the variable abbreviation):

\begin{table}[H]
\centering
\begin{NiceTabular}{lccccccccccc} 
\toprule
\textbf{Variable} & z & t & u & v & q & t2m & MSLP & u10 & v10 & TP6 & TP24 \\
\cmidrule(lr){2-12}
\textbf{Weight} & 1.0 & 1.0 & 0.5 & 0.5 & 0.1 & 1.0 & 0.1 & 0.1 & 0.1 & 0.05 & 0.05 \\
\bottomrule
\end{NiceTabular}
\caption{Variable-based weight. \label{tab:variable weight}}
\end{table}

We didn't perform search over the best combination of variable-based weights and level-based weights. We find uniform level-based weights perform almost similar to current level-based weights in preliminary experiments. The variable weights we've used is similar to the variable weights used in GraphCast\citep{graphcast2023science}. Recent work Stormer \citep{nguyen2023stormer} showcase that using variable-based weight is beneficial for variables that are assigned higher loss weight like 2m temperature, geopotential.

\subsection{Metrics}
\label{appendix metric definition}

The following metrics are defined on a surface variable or a upper-air variable at a specific level.
\paragraph{Root mean squared error (RMSE)}
The RMSE for time step $t$ is defined as:
\begin{equation}
    \label{eq:rmse}
    \text{RMSE}_t(\hat{X}, X) = \sqrt{
    \frac{1}{N_{\text{lat}}N_\text{lon}}
    \sum_{i=1}^{N_{\text{lat}}}
    \sum_{j=1}^{N_{\text{lon}}}
    w(i)(\hat{X}_{tij}-X_{tij})
    }.
\end{equation}

\paragraph{Anomaly correlation coefficient (ACC)} The ACC is defined as the Pearson correlation coefficient of the anomalies with respect to the
climatology:
\begin{align}
    \label{eq:acc}
    &\text{Anomaly:} \quad \hat{X}'_{tij} = \hat{X}_{tij} - C_{\tilde{t}ij}, \quad  X'_{tij} = \hat{X}_{tij} - C_{\tilde{t}ij}, \\
    &\text{ACC}_t(\hat{X}', X') = \frac{
     \sum_{i=1}^{N_{\text{lat}}}
    \sum_{j=1}^{N_{\text{lon}}}w(i)\hat{X}'_{tij}X'_{tij}}
    {\sqrt{ 
    \left(\sum_{i=1}^{N_{\text{lat}}}
    \sum_{j=1}^{N_{\text{lon}}}w(i)\hat{X'}^2_{tij}\right)
     \left(\sum_{i=1}^{N_{\text{lat}}}
    \sum_{j=1}^{N_{\text{lon}}}w(i)X^2_{tij}\right)}},
\end{align}
where we use the climatology statistics $C$ provided in WeatherBench2 \citep{rasp2023weatherbench2} (computed on ERA5 data from $1990$ to $2019$), $\tilde{t}$ denotes the day of year and time of day of the corresponding time index $t$.

\paragraph{Bias} The bias is defined as the point-wise difference between prediction and ground truth:
\begin{equation}
     \label{eq:bias}
    \text{Bias}_t(\hat{X}, X) = \hat{X}_{tij} - X_{tij}.
\end{equation}%
\section{Comparison against standard attention}
\label{appendix attention comparison}
\begin{figure}[H]
    \centering
    \includegraphics[width=0.9\linewidth]{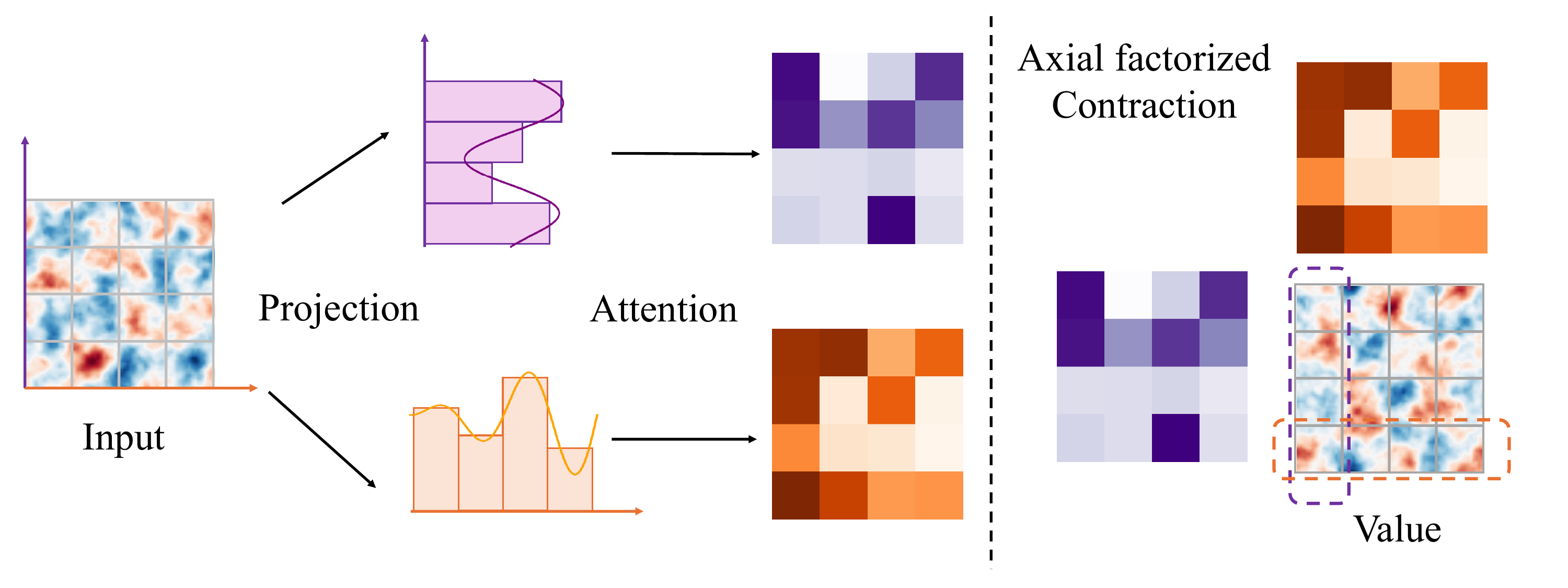}
    \caption{Axial factorized attention. The multi-dimensional spatial structure of value is preserved, each axial attention kernel matrix's size is $S_m^2$ for $n$-dimensional grid with total size $ S=S_1\times S_2\times \hdots \times S_n$. }
    \label{fig:diagram factorized attention}
    \vspace{2mm}
    \includegraphics[width=0.9\linewidth]{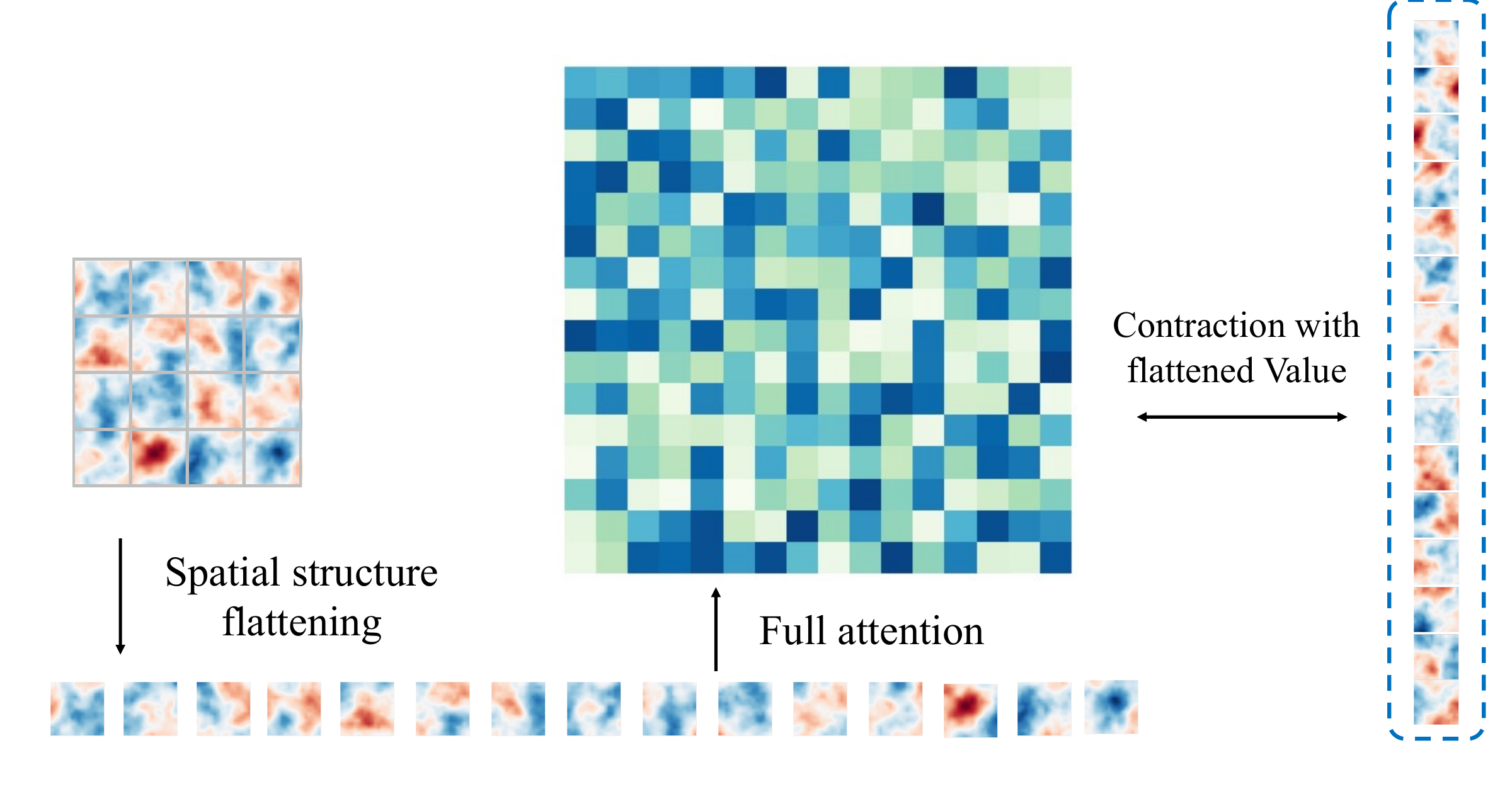}
    \caption{Standard attention for sequence. The multi-dimensional spatial structure is not considered explicitly during the attention (it can be implicitly encoded via positional encoding). The full attention kernel matrix's size is quadratic to the total grid size $S$.}
    \label{fig:diagram attention}
\end{figure}

Two illustrative diagrams of factorized attention and standard attention are shown above (Figure \ref{fig:diagram factorized attention} and \ref{fig:diagram attention}). Instandard attention, the spatial structure is not preserved during the contraction of value and attention kernel. Compared to axial Transformer \citep{ho2020axial}, the axial attention kernel in CaFA is still dependent on global information while in axial Transformer its context is limited to that specific row/column. 

\begin{figure}[H]
    \centering
    \vspace{-2mm}
    \begin{subfigure}{\textwidth}
    \centering
    \includegraphics[width=0.6\textwidth]{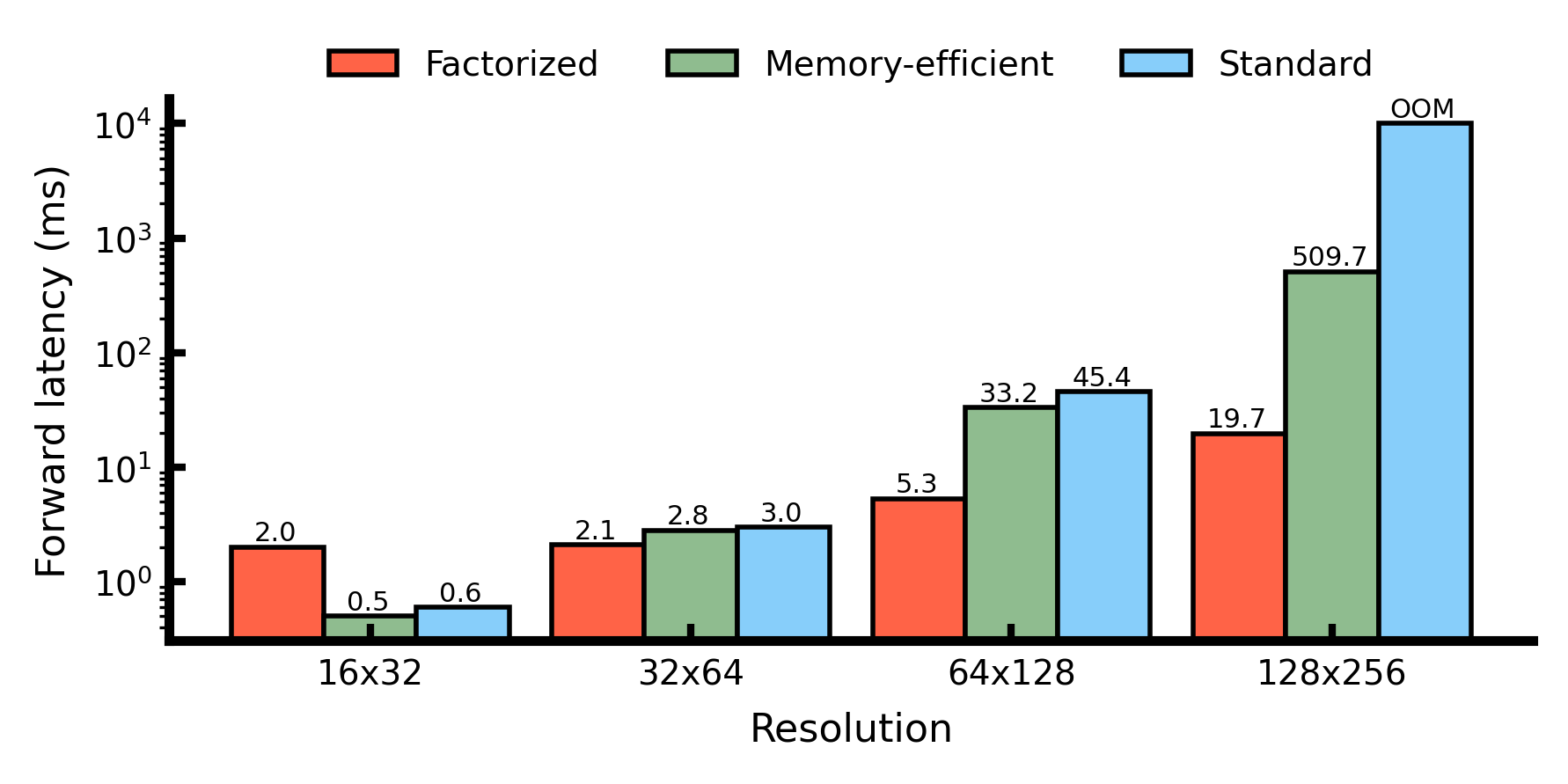}
    \caption{Forward latency}
    \label{fig:latency}
    \end{subfigure}
    \begin{subfigure}{\textwidth}
        \centering
        \includegraphics[width=0.6\textwidth]{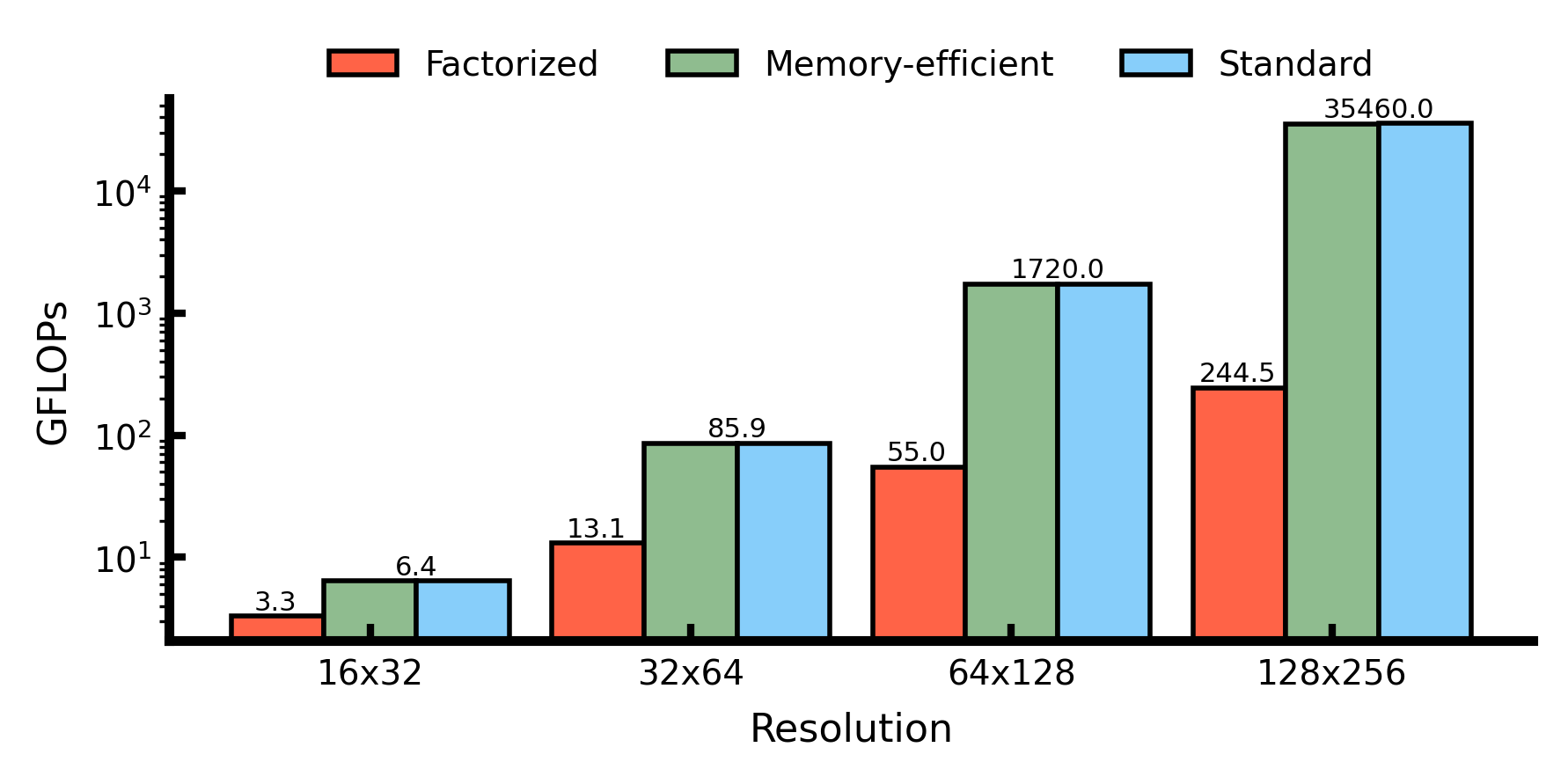}
    \caption{Theoretical FLOPs}
    \label{fig:flops}
    \end{subfigure}
    \begin{subfigure}{\textwidth}
        \centering
        \includegraphics[width=0.6\textwidth]{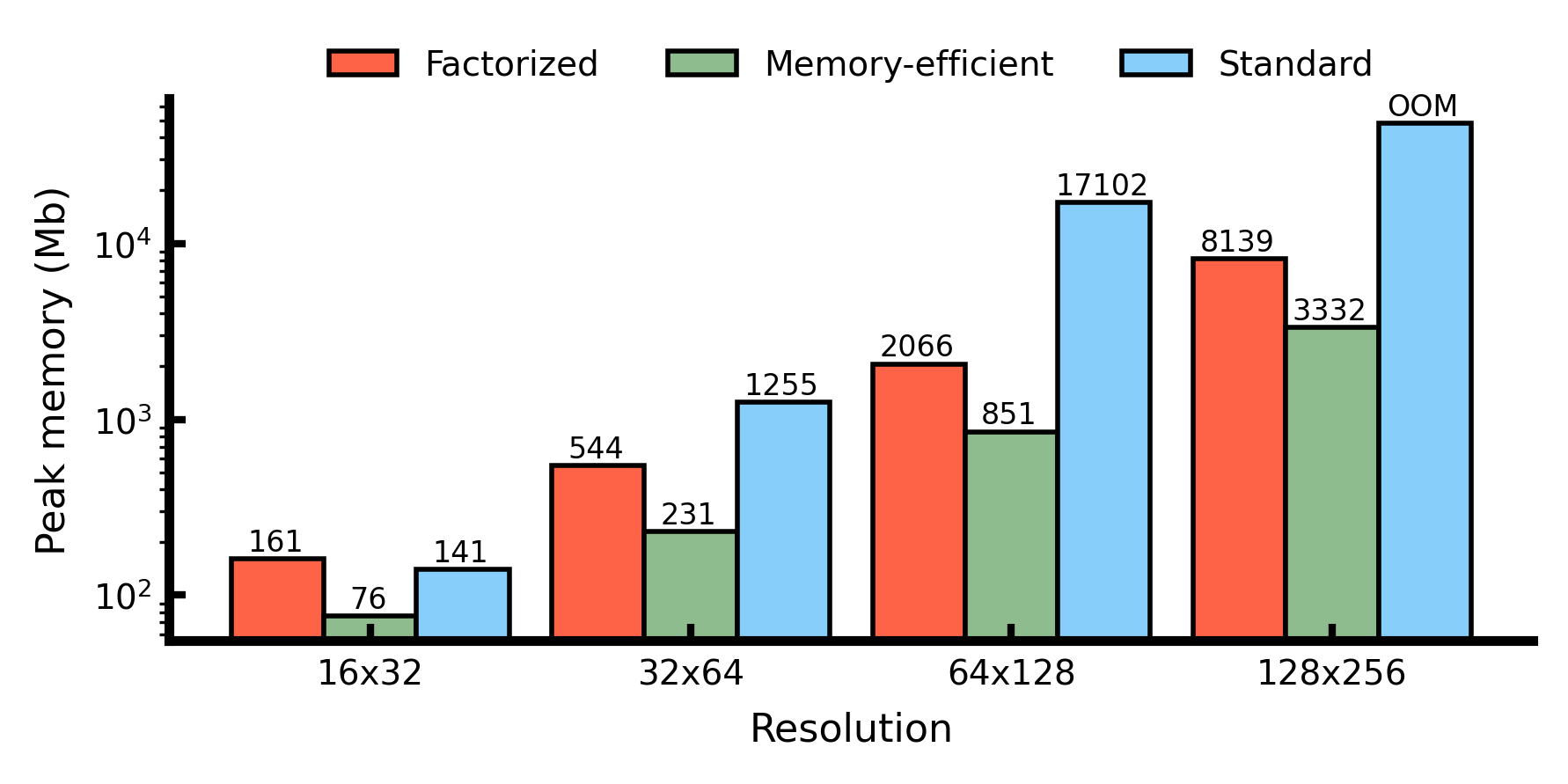}
    \caption{Peak memory usage}
    \label{fig:memory}
    \end{subfigure}
    \vspace{-2mm}
    \caption{\label{fig:computational benchmark}Computational benchmark between factorized attention and different implementation of the standard softmax attention \citep{attention2017nips}. The number of attention head is 16, each attention head's dimension is 128, batch size is set to 1, input and output channel dimension is 512. The benchmark is carried out on A6000 with CUDA 12, \texttt{float32} precision is used. The runtime statistics are profiled using DeepSpeed. The factorized attention layer contains following learnable modules that are not in standard attention layer: distance encoding (Equation \eqref{eq:distance encoding}) and projection layer (Equation \eqref{eq:attn projection}).}
        \vspace{-2mm}
\end{figure}

We compare factorized attention to two different implementations of standard attention in PyTorch's attention function: \verb|torch.nn.functional.scaled_dot_product_attention|
- a standard one without memory optimization and another one with memory optimization \citep{rabe2021memefficient} \footnote {FlashAttention \citep{flashattention2022nips} is currently not available for \texttt{float32} computation on A6000 in PyTorch 2.1.}. The results are shown in Figure \ref{fig:latency}, \ref{fig:flops}, \ref{fig:memory}. We observe that factorized attention has significantly better scaling efficiency in terms of runtime and FLOPs. At the lowest resolution, the runtime of factorized attention is larger than standard attention. This is because in factorized attention layer the input is first processed by two projection layer and then sent to compute the axial attention kernels, with all computation done sequentially  \footnote{For projection and attention kernel's computation, each axis's computation are independent to each other and the memory cost is small, so potentially they can be optimized to run in parallel.}. Compared to memory efficient attention, factorized attention has larger memory footprint at current stage. The memory cost of factorized attention can also be optimized by adopting a similar chunking strategy in \citet{rabe2021memefficient} during the contraction of value and attention kernels.

\section{Further results}
\label{appendix more results}

In this section we provide more quantitative and qualitative analysis of model's prediction capability.

\begin{figure}[H]
    \centering
    \includegraphics[width=\textwidth]{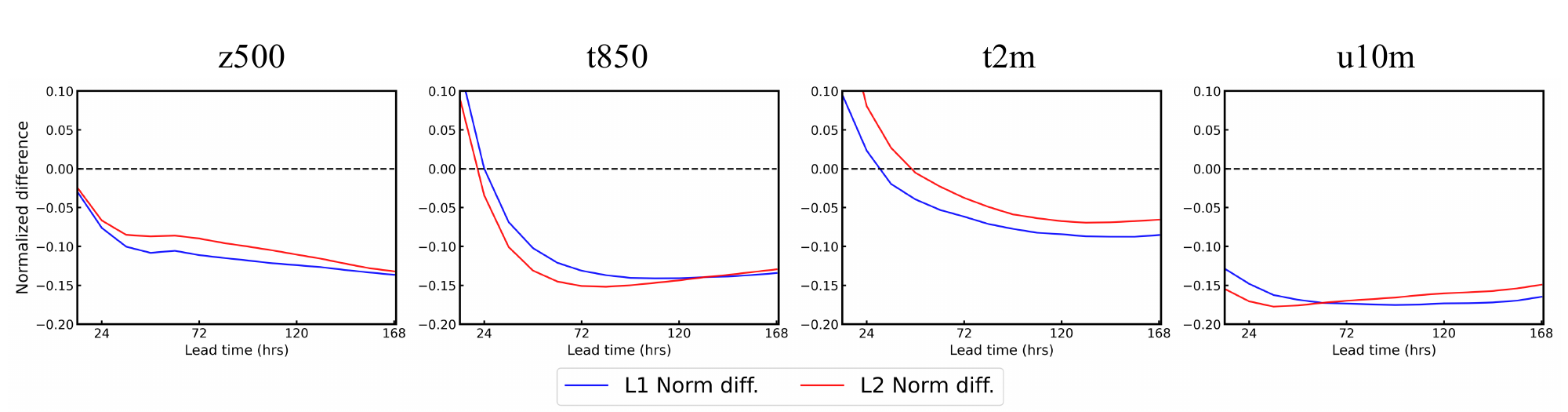}
    \caption{Relative $L1$ and $L2$ error difference with respect to IFS HRES of 4 selected key variables.}
    \label{fig:l1-l2-diff}
\end{figure}

We analyze how the model perform in terms of $L1$ and $L2$ (RMSE) loss. The normalized difference with respect to IFS HRES is shown in Figure \ref{fig:l1-l2-diff}. We observe that the model performs relatively better in terms of $L1$ norm in the long run, which is possibly because that the model is trained on $L1$ norm and penalizes outlier less than $L2$ norm. We also compare models trained with different time step sizes (lead time 6 hour vs 12 hour). We find that model with larger time step size performs better in terms of forecasting accuracy beyond 7 days but performs worse in shorter range, while the ACC of both model variants fall below $0.6$ on day 10.

\begin{figure}[H]
    \centering
    \includegraphics[width=0.85\textwidth]{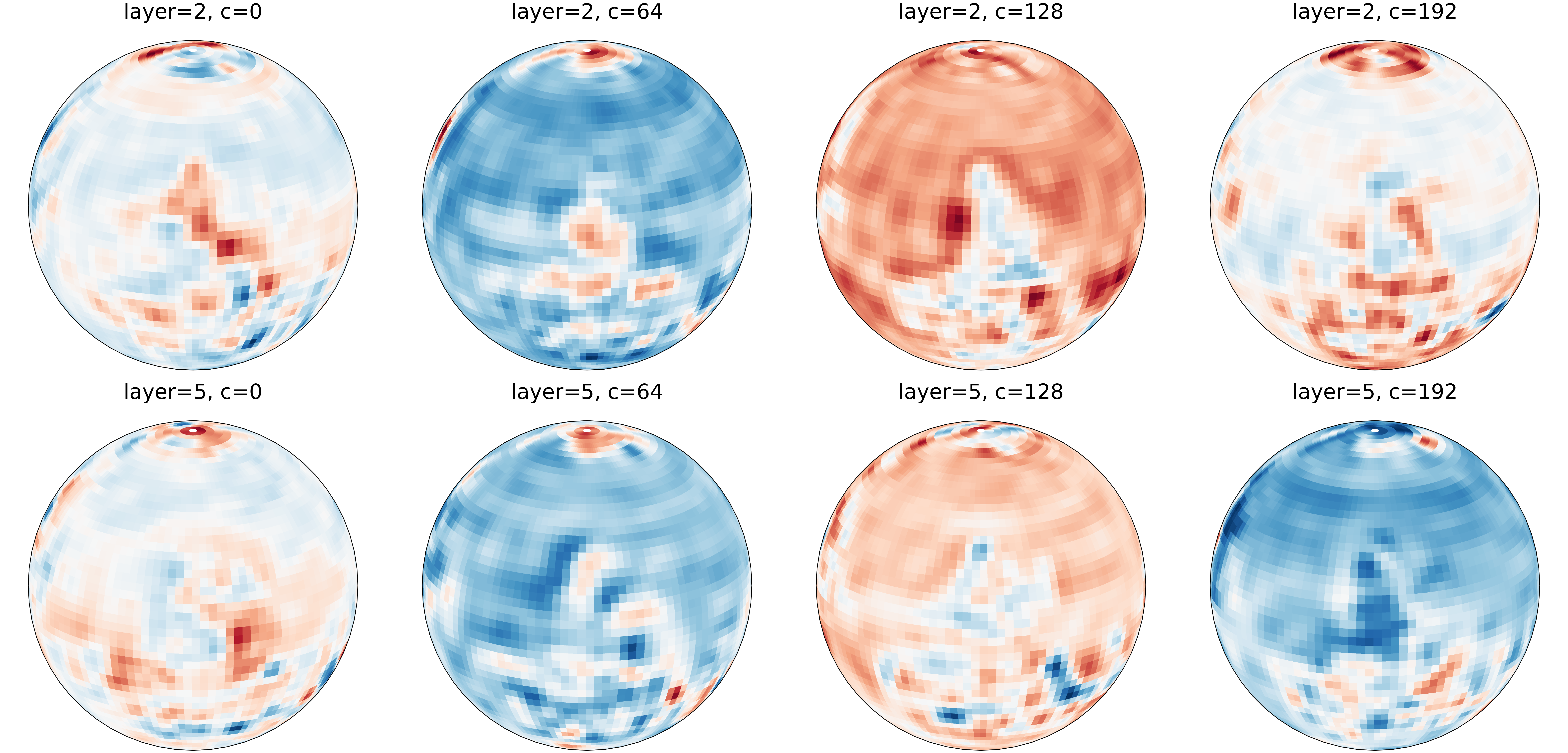}
    \caption{Example visualization of learned spherical harmonic based positional encoding. "Layer" denotes their corresponding attention layer number, "c" denotes channel number.}
    \label{fig:spherical pe}
\end{figure}

\begin{figure}[H]
    \centering
    \includegraphics[width=0.90\textwidth]{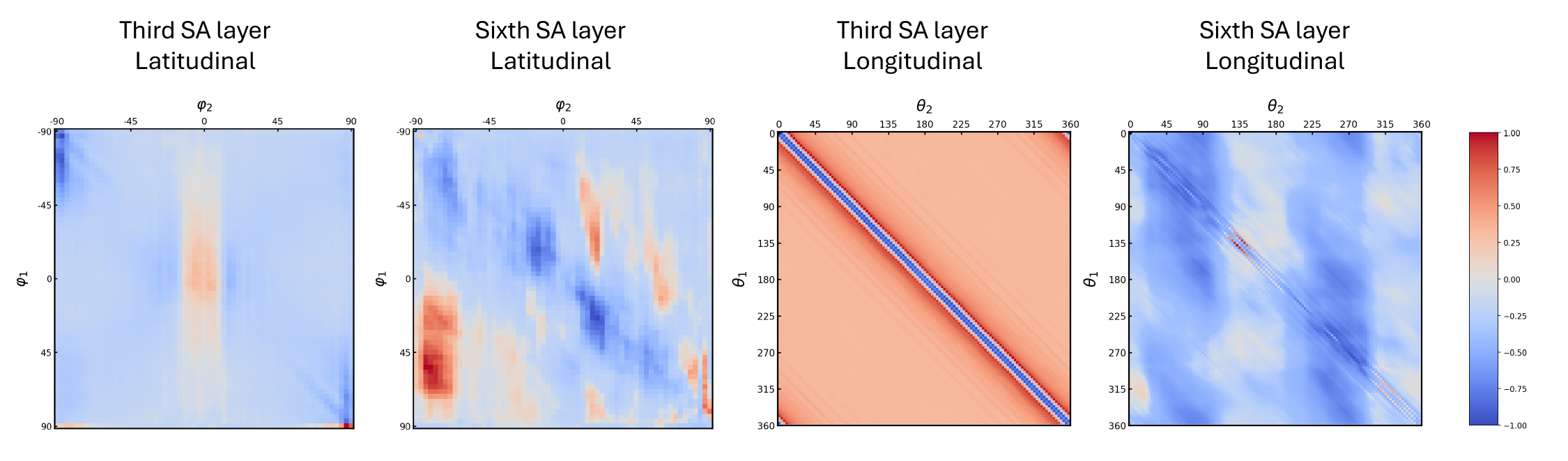}
    \caption{Example visualization of learned self-attention kernel along different axes. The kernel matrices are selected from random batch and head. For better clarity all the kernel matrices shown are normalized such that all the elements fall into the range of $[-1, 1]$.}
    \label{fig:attention kernel visualization}
\end{figure}

\begin{figure}[H]
    \centering
    \includegraphics[width=0.90\textwidth]{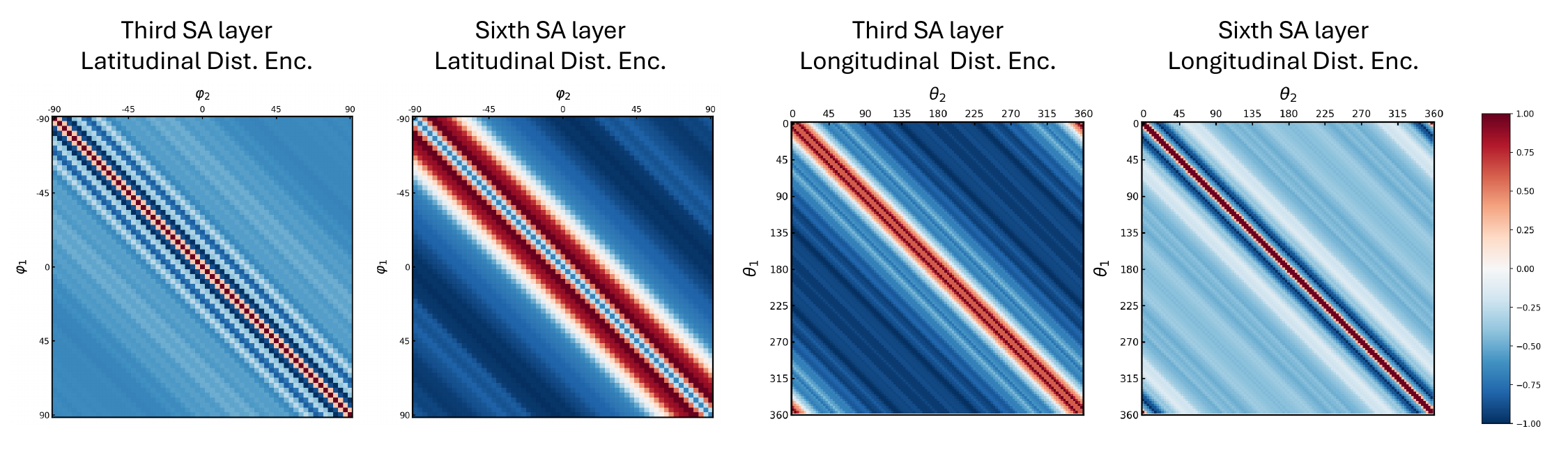}
    \caption{Example visualization of learned distance encoding along different axes. The distance encoding are selected from random batch, head and channel. For better clarity all the kernel matrices shown are normalized such that all the elements fall into the range of $[-1, 1]$.}
    \label{fig:distance encoding visualization}
\end{figure}

Example visualization of the learned positional encoding, distance encoding and attention kernels are shown in Figure \ref{fig:spherical pe}, \ref{fig:attention kernel visualization}, \ref{fig:distance encoding visualization}. Interestingly, the attention kernel at earlier layer exhibit sharper patterns than deeper layer, where attention scores are mostly concentrated around the diagonal.

\begin{figure}[H]
    \centering
    \includegraphics[width=\textwidth]{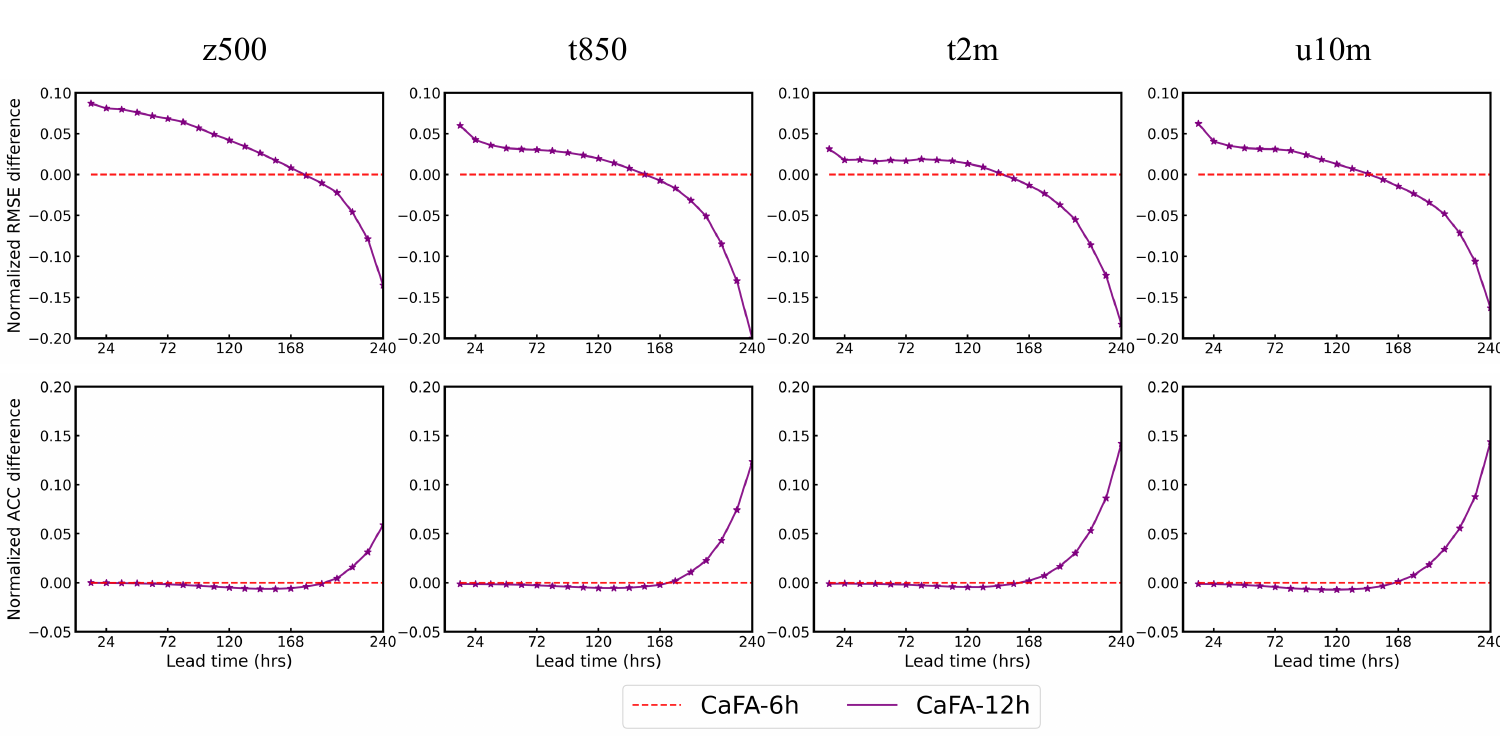}
    \caption{Normalized RMSE difference and ACC difference comparison between CaFA trained with $6$ hour interval and CaFA trained with $12$ hour interval. Negative RMSE difference and positive ACC difference indicates better performance. CaFA trained with $12$ hour interval uses less gradient steps for the second stage (36k) and fewer rollout steps (up to 16).}
    \label{fig:6hvs12h}
\end{figure}

\begin{figure}[H]
    \centering
    \begin{subfigure}{\textwidth}
    \includegraphics[width=0.94\textwidth]{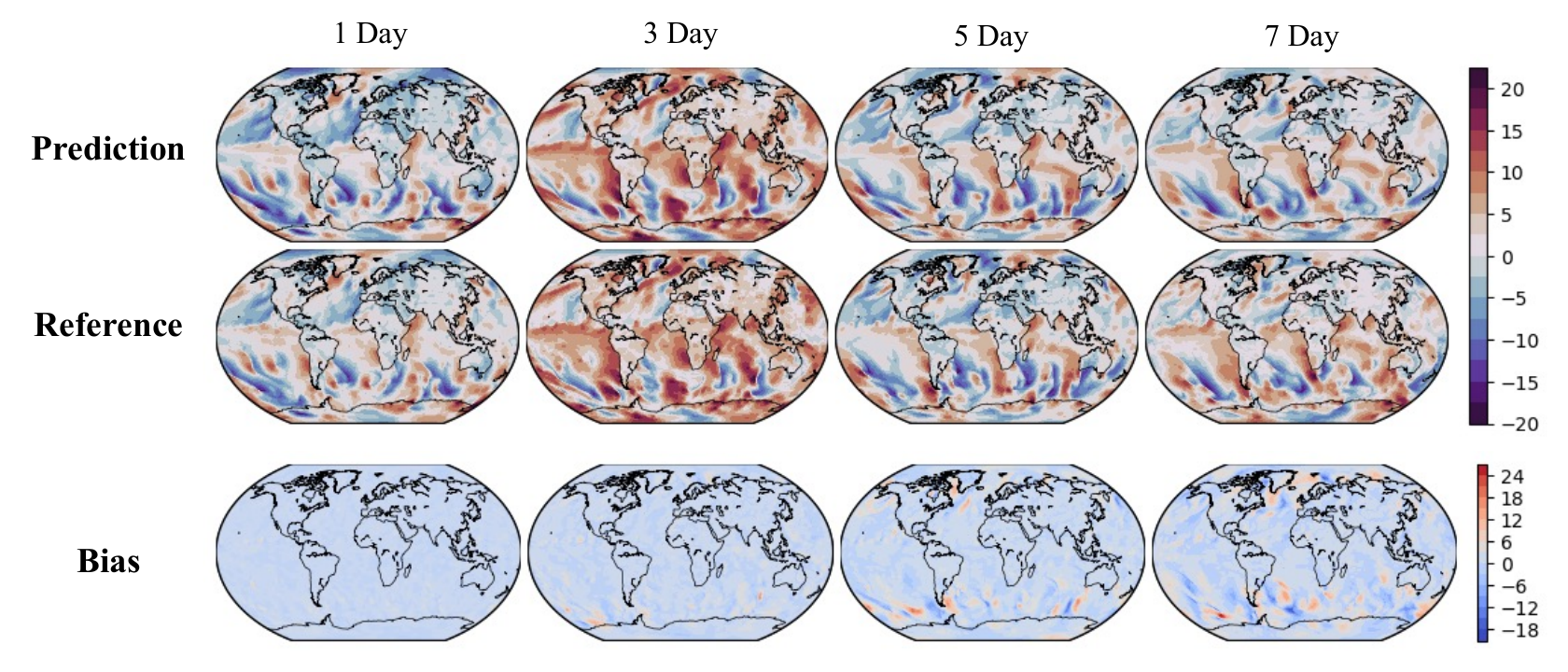}
    \vspace{-1mm}
    \caption{$v10$m example rollout visualization}
    \end{subfigure}  
    \begin{subfigure}{\textwidth}
    \includegraphics[width=0.94\textwidth]{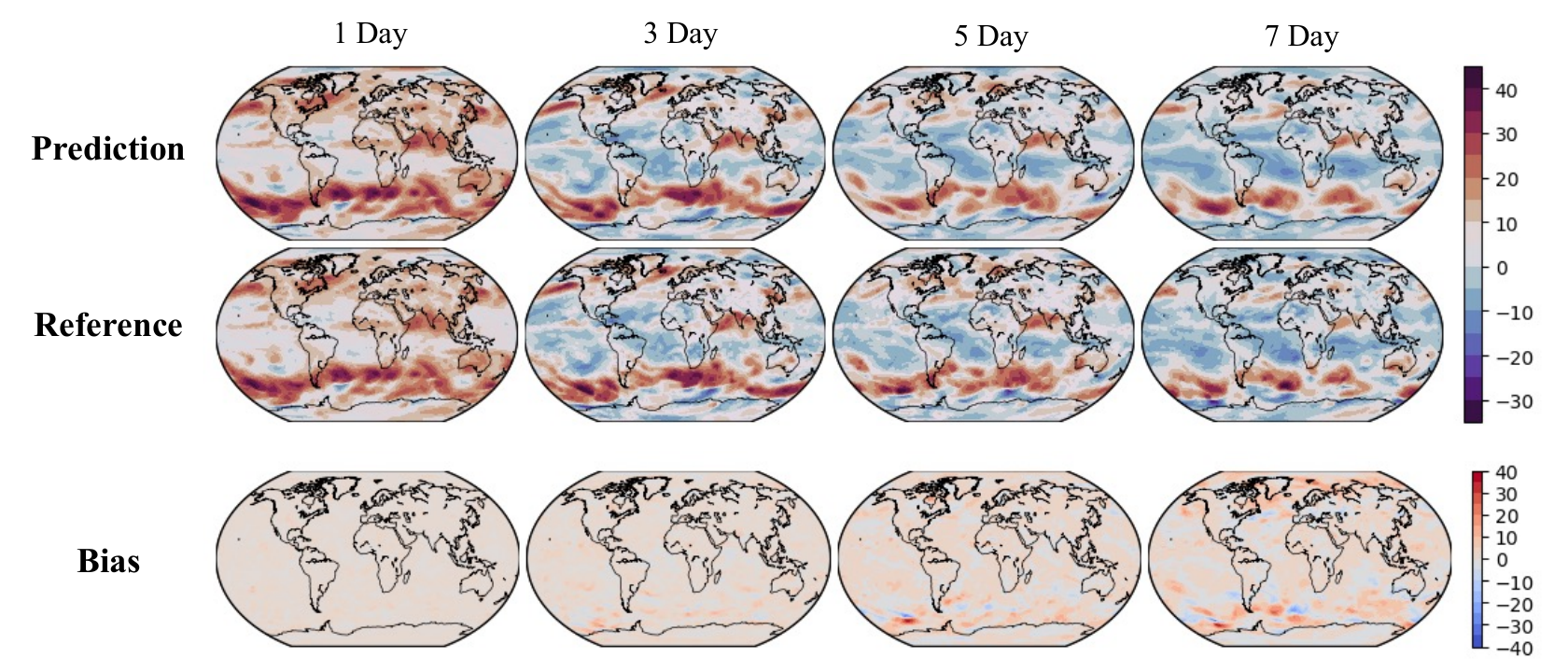}
    \vspace{-1mm}
    \caption{$u850$ example rollout visualization}
    \end{subfigure}
    \vspace{-1mm}
    \begin{subfigure}{\textwidth}
    \includegraphics[width=0.94\textwidth]{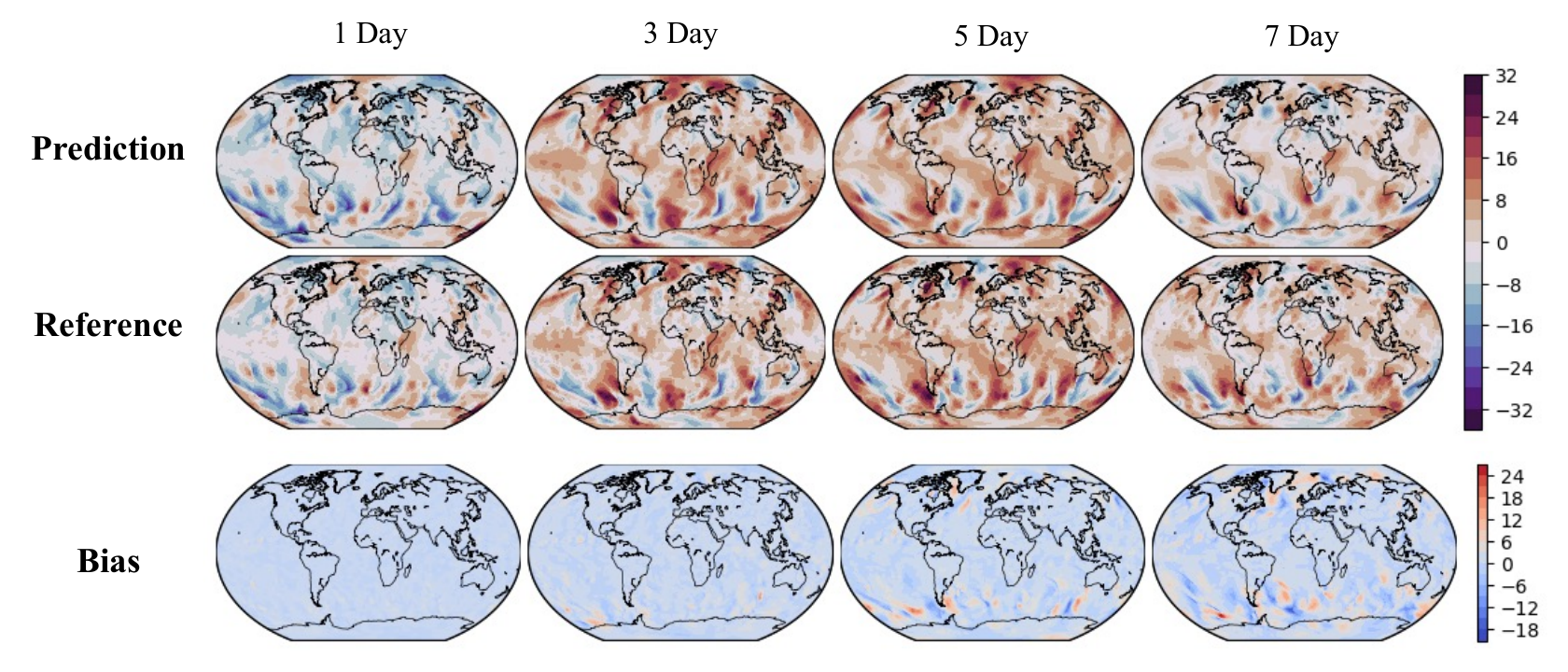}
    \vspace{-1mm}
    \caption{$v850$m example rollout visualization}
    \end{subfigure}
\end{figure}%
\begin{figure}[H]\ContinuedFloat 
    \medskip
    \begin{subfigure}{\textwidth}
    \includegraphics[width=0.94\textwidth]{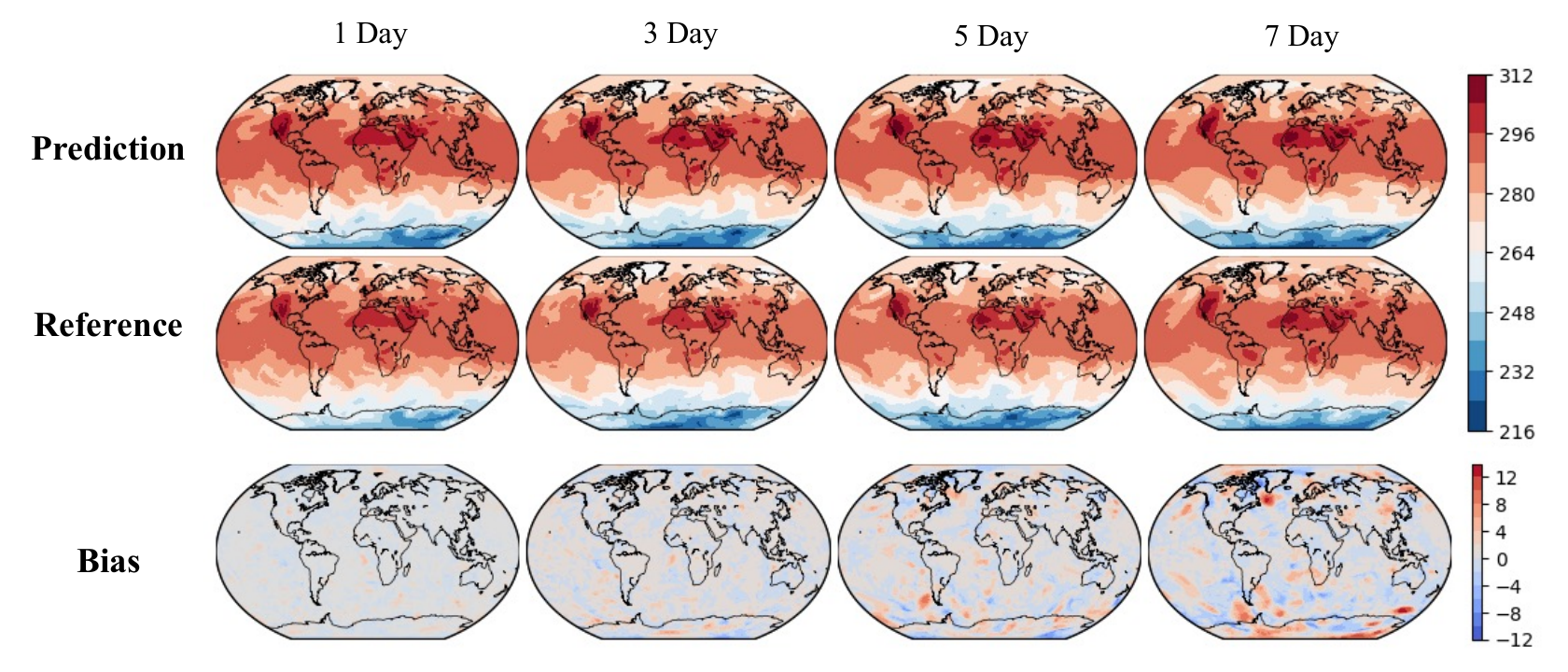}
    \vspace{-1mm}
    \caption{$t850$ example rollout visualization}
    \end{subfigure}
    \begin{subfigure}{\textwidth}
    \includegraphics[width=0.94\textwidth]{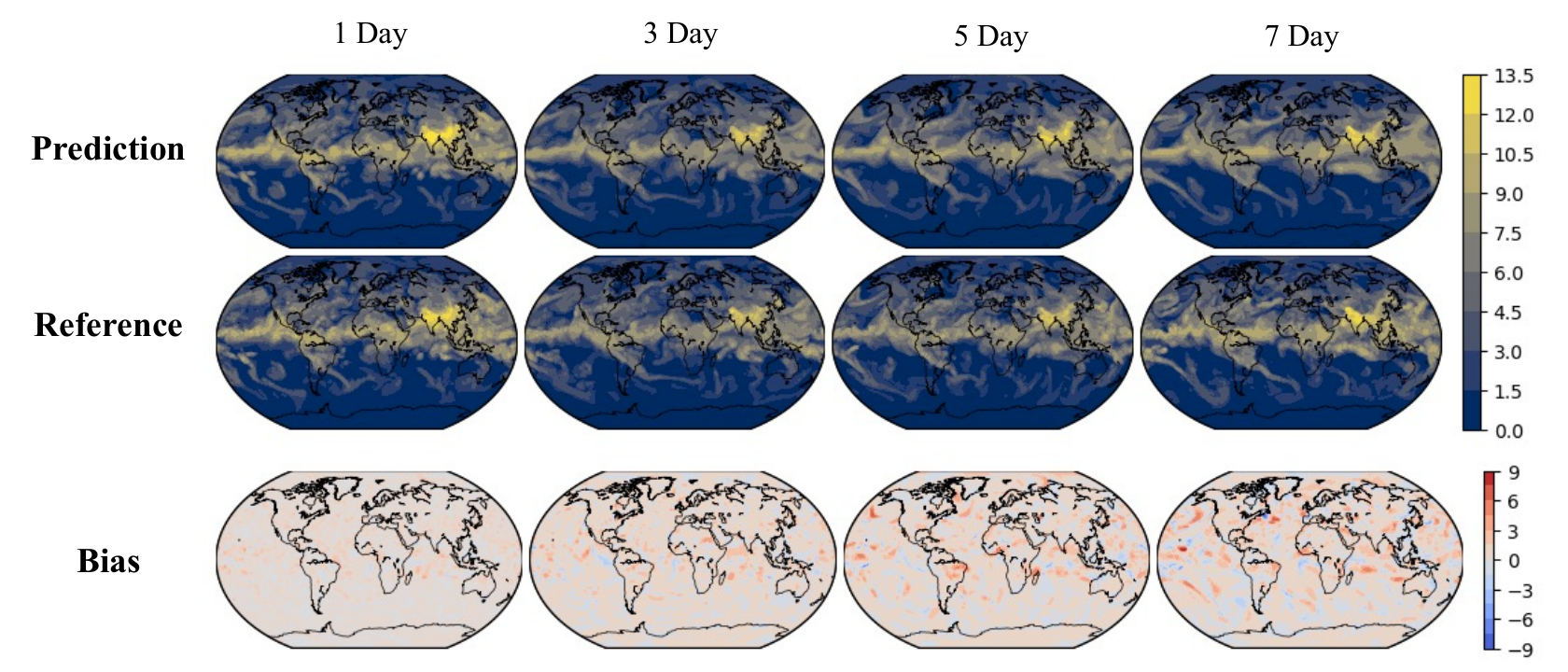}
    \vspace{-1mm}
    \caption{$q700$m example rollout visualization}
    \end{subfigure}
    \begin{subfigure}{\textwidth}
    \includegraphics[width=0.94\textwidth]{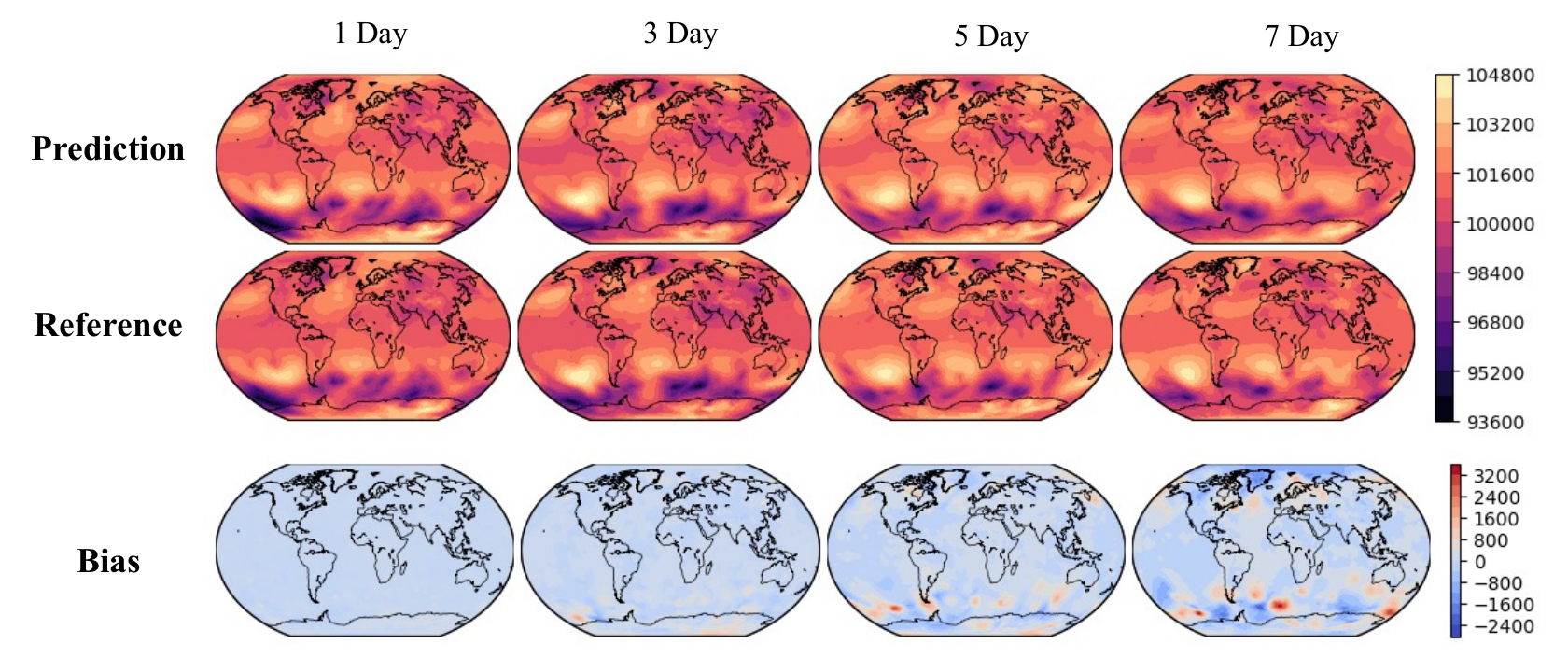}
    \vspace{-2mm}
    \caption{MSLP example rollout visualization}
    \end{subfigure}%
    \vspace{-3mm}
    \caption{\label{fig:more qualitative visualization} More example rollout visualization of model's prediction versus reference ERA5 reanalysis data at different lead times. The initialization time is 00:00 UTC on August 11, 2020. }
    \vspace{-10mm}
\end{figure}
\end{document}